%% file: main.tex
\documentclass{article}

\usepackage[nonatbib]{neurips_2020}
\usepackage{pslatex}
\usepackage{hyperref}
\usepackage{lineno}
\usepackage{xcolor}
\usepackage{bookmark}
\usepackage{graphicx}
\usepackage{tikz}
\usepackage{amsmath}
\usepackage{amssymb}
\usepackage{amsthm}
\usepackage{mathtools}
\usepackage{placeins}
\usepackage{bbm}
\usepackage[capitalize,noabbrev]{cleveref}
\usepackage{enumitem}
\usepackage{booktabs}
\usepackage{wrapfig}
\usepackage{caption}
\usepackage{soul}
\usepackage{bbold}
\usepackage{multirow}
\usepackage{makecell}
\usepackage{array}
\usepackage{subcaption}
\usepackage{todonotes}
\usepackage{changes}
\usepackage{setspace}
\usepackage{float}
\hypersetup{colorlinks=true, linkcolor=red!90!black, citecolor=green!50!black, urlcolor=blue!90!black}
\usepackage[style=authoryear,backend=biber,natbib,giveninits=true,uniquename=init,sorting=nyt,maxcitenames=1,mincitenames=1,minbibnames=6,maxbibnames=6,hyperref=true]{biblatex}
\makeatletter
\renewbibmacro*{cite}{%
    \printtext[bibhyperref]{%
        \iffieldundef{shorthand}
        {\ifboolexpr{test {\ifnameundef{labelname}} or test {\iffieldundef{labelyear}}}
            {\usebibmacro{cite:label}%
            \setunit{\printdelim{nonameyeardelim}}}
            {\printnames{labelname}%
            \setunit{\printdelim{nameyeardelim}}}%
        \usebibmacro{cite:labeldate+extradate}}
{\usebibmacro{cite:shorthand}}}}
\makeatother

\title{
    Metric Learning Encoding Models: A Multivariate Framework for Interpreting Neural Representations
}

\begin{document}



\maketitle

\vspace{-2cm}

\begin{center}
    \large \textbf{Louis~Jalouzot$^{1,2}$, Christophe~Pallier$^{1}$, Emmanuel~Chemla$^{2,3\ast}$, Yair~Lakretz$^{2\ast}$}\\
    \vspace{.25cm}
    \normalsize $^{1}$UNICOG, CNRS, INSERM, CEA, Paris-Saclay University\\
    $^{2}$LSCP, EHESS, ENS, CNRS, PSL University\\
    $^{3}$Earth Species Project\\
    $^\ast$ Equal contribution\\
    \vspace{.25cm}
    \small \textbf{Correspondance:} \href{mailto:jalouzot.louis@gmail.com}{jalouzot.louis@gmail.com}
\end{center}


\begin{abstract}
    Understanding how explicit theoretical features are encoded in opaque neural systems is a central challenge now common to neuroscience and AI.
    We introduce Metric Learning Encoding Models (MLEMs) to address this challenge most directly as a metric learning problem: we fit the distance in the space of theoretical features to match the distance in neural space.
    Our framework improves on univariate encoding and decoding methods by building on second-order isomorphism methods, such as Representational Similarity Analysis, and extends them by learning a metric that efficiently models feature as well as interactions between them.
    The effectiveness of MLEM is validated through two sets of simulations.
    First, MLEMs recover ground-truth importance features in synthetic datasets better than state-of-the-art methods, such as Feature Reweighted RSA (FR-RSA).
    Second, we deploy MLEMs on real language data, where they show stronger robustness to noise in calculating the importance of linguistic features (gender, tense, etc.).
    MLEMs are applicable to any domains where theoretical features can be identified, such as language, vision, audition, etc.
    We release optimized code applicable to measure feature importance in the representations of any artificial neural networks or empirical neural data at \href{https://github.com/LouisJalouzot/MLEM}{https://github.com/LouisJalouzot/MLEM}.
\end{abstract}



\input{introduction.tex}
\input{methods.tex}
\input{results.tex}
\input{discussion.tex}

\clearpage

\section*{Acknowledgements}

This work was supported by grants ComCogMean (ANR-23-CE28-0016), FrontCog (ANR-17-EURE-0017), Orisem (ERC Grant Agreement N° 788077), and ANR-10-IDEX-0001-02 and was performed using HPC resources from GENCI-IDRIS (Grant 2024-AD011016055).

\printbibliography

\clearpage

\appendix

\begin{center}
    \LARGE{\textbf{Supplementary Information}}
\end{center}

\setcounter{figure}{0}
\setcounter{table}{0}
\renewcommand{\thefigure}{S\arabic{figure}}
\renewcommand{\thetable}{S\arabic{table}}

\input{appendix.tex}

\end{document}

%% file: introduction.tex
\begin{figure}[!ht]
    \centering
    \includegraphics[width=.77\textwidth]{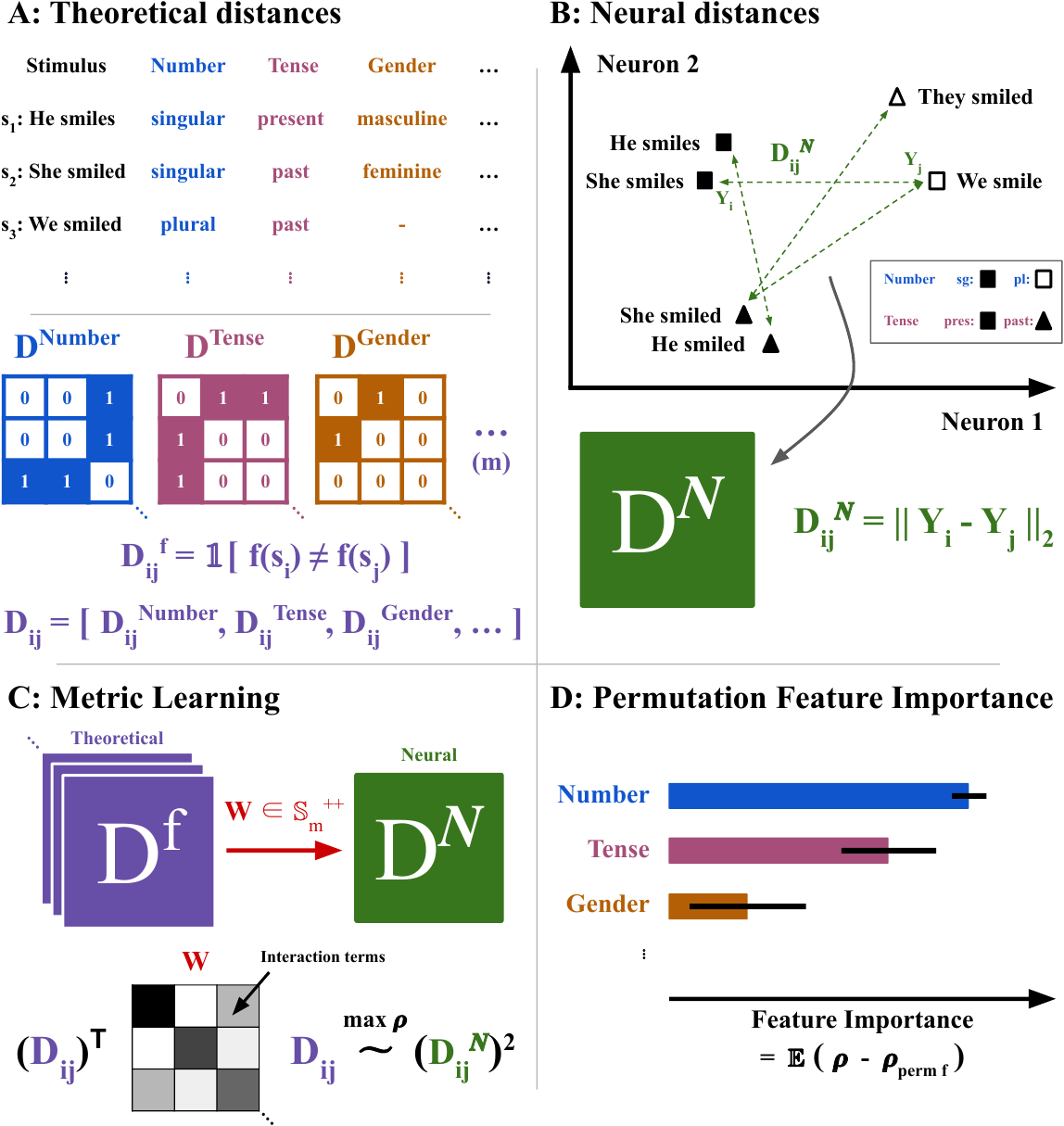}
    \vspace{0.5cm}
    \caption{
        \textbf{Overview of the Metric Learning Encoding Models (MLEMs) Approach:} an example from natural language\\
        \textbf{A:}
        Consider a set of $n$ stimuli (rows; sentences) and $m$ theoretical features (columns; grammatical number, tense, gender, ...) along with their values for each stimulus.
        In this example, the features are categorical and \textit{feature distance} are defined as whether a pair of stimuli ($s_i, s_j$) share a given feature $f_k$: $\mathbb{1}\left[f_k(s_i) \neq f_k(s_j)\right]$.
        Accordingly, a Representational Dissimilarity Matrix (RDM) is computed for each of the $m$ features ($D^\text{Number}, D^\text{Tense}, D^\text{Gender}, ... \in \mathbb{R}_+^{n \times n}$).
        Elements of the matrix thus indicate whether pairs of stimuli share a given feature.
        Finally, for each pair of stimuli, let $D^F_{ij} \in \mathbb{R}^m$ be a vector, composed of the $(i, j)$ elements from all RDMs (bottom panel).
        $D^F_{ij}$ summarizes the \textit{theoretical distance} between two stimuli, with respect to all features.\\
        \textbf{B:}
        On the other hand, consider $n$ high-dimensional neural representations of the stimuli (e.g. embeddings of LLMs or fMRI images, here represented in 2D).
        From these representations, one can compute an RDM of \textit{neural distances} $D^N$ (e.g., based on Euclidean distance).\\
        \textbf{C:}
        The core metric learning component of MLEM involves learning a weighted norm of the feature distance vectors $\|D^F_{ij}\|_W$ to approximate the empirical neural distances $D_{ij}^N$.
        This approximation is optimized to maximize the Spearman correlation $\rho$ between the theoretical and neural distances.
        The norm is parametrized with a learned Symmetric Positive Definite (SPD) matrix $W\in\mathbb{S}^{++}_m$ so as to be a valid metric (distance function).
        Crucially, the off-diagonal coefficients of $W$ allow for direct modeling of interactions between features (e.g., between grammatical number and gender) since they may have a combined effect on neural representations.\\
        \textbf{D:}
        Permutation Feature Importance is applied to a fitted MLEM model to assess the importance of each feature and their interactions in predicting neural distances.
        In practice, the importance is the average decrease in Spearman correlation $\rho$ when the feature values are permuted.
    }
    \label{fig:mlem}
\end{figure}

\section{Introduction}

A central goal in both neuroscience and Artificial Intelligence is to understand how complex, high-dimensional systems like the brain and deep neural networks represent and process information \citep{belinkov_analysis_2019, pohl_clarifying_2024}.
These systems excel at complex tasks such as natural language processing but their internal mechanisms are still not well understood and for this reason are often called ``black boxes''.
High-dimensional recordings of neural activity can be obtained from both biological and artificial systems, but interpreting these complex data remains a major challenge.

A powerful strategy for interpreting these systems is to decompose those internal representations into a set of discrete, theoretically-grounded features.
For instance, in the domain of language, it is natural to ask how different aspects of language, such as syntax \citep{pallier_cortical_2011, hewitt_structural_2019,diego-simon_polar_2024}, semantics \citep{huth_natural_2016, li_geometry_2024}, or morphology \citep{crepaldi_probabilistic_2019} are encoded in neural recordings from the human brain or in the activation patterns of a Large Language Model (LLM).
Similarly, in vision, one might investigate how features like object category, shape, or texture are represented in the visual cortex or in a convolutional neural network.

Two general approaches to study the neural encoding of theoretical features can be discerned in the literature: \textit{decoding} and \textit{encoding} models \citep{kriegeskorte_interpreting_2019}.
\textbf{Decoding models} (or ``probes") attempt to predict features from neural activity \citep{hupkes_visualisation_2018, tenney_what_2019, arps_probing_2022}.
High performance of a decoding model demonstrates that feature-related information is present but this approach is susceptible to confounds.
Indeed a feature may be decodable simply because it correlates with another, truly encoded feature \citep{hewitt_designing_2019, belinkov_probing_2022, kumar_probing_2022}.
For example, a probe trained to decode tense in a sentence from some representations might succeed not because tense is explicitly encoded in the representations, but because, e.g., future tense correlates with the superficial presence of ``will'' in the sentence, which is all the model needs to encode for success without any further level of abstraction.
\textbf{Encoding models}, conversely, predict neural activity from features \citep{wehbe_simultaneously_2014, schrimpf_neural_2020, caucheteux_longrange_2021, caucheteux_brains_2022, pasquiou_neural_2022, oota_joint_2022, pasquiou_informationrestricted_2023, hosseini_artificial_2024}.
Those models can be fitted on multiple features and confounds and allow for delineating their contribution.
However, unlike decoding models, they are often univariate, fitting each neural unit (e.g., neuron or voxel) independently.
Therefore, they cannot model distributed representations encoded across populations of units, which is a critical characteristic of neural computation \citep{georgopoulos_neuronal_1986,rumelhart_parallel_1986}.

Here, we introduce a simple approach, which preserves the best from both worlds, by extending encoding methods to the multivariate case within a metric learning framework \citep{kulis_metric_2013}.
In a nutshell, the method learns an optimal metric over a space of theoretical features to best match the geometry of the neural representation space \citep{goldstein_alignment_2024, brown_topology_2024}.
We show that previous methods, such as Representational Similarity Analysis (RSA; \cite{kriegeskorte_representational_2008}) and its feature-reweighted variant (FR-RSA, \cite{jozwik_deep_2017, storrs_diverse_2021, xu_weighted_2021, kaniuth_featurereweighted_2022}) can be derived as \textit{special cases} of MLEMs.
We call the method Metric Learning Encoding Models, or MLEMs for short.
A key innovation of MLEM is its ability to model interactions between features by design, a critical requirement for understanding complex domains like language where features are rarely processed in isolation and are often entangled in a single representation vector \citep{huang_disentangling_2021, kumar_shared_2022}.
For instance, in English, the marking of a verb with an ``-s" suffix depends on the interaction between two features of its subject: its number (singular) and its person (third).
Hence a neural representation of the verb has to jointly encode both features to be used downstream to produce the correct morphological form.


The approach is validated in two settings.
First, using simulated data with a known ground-truth, we show that MLEM recovers underlying feature weights more accurately than a strong baseline, FR-RSA.
Second, when applied to representations from an LLM processing controlled syntactic stimuli, MLEM proves more robust to noise and converges faster, while identifying similar feature importance profiles to FR-RSA.
These improvements in robustness and data efficiency are especially useful for real-world applications, such as analyzing feature encoding in brain data which is often noisy and scarce.


The present study focuses on text stimuli and LLM representations, but the MLEM framework is modality-agnostic.
As a second-order isomorphism method, it can be applied to any domain where stimuli can be described by a set of theoretical features and measured in a high-dimensional representation space.
This includes any experiment involving visual or auditory stimuli and neural representations can come from various neuroimaging modalities (fMRI, MEG, ECoG, etc.) or even from intermediate layers of artificial neural networks.
This work thus provides a versatile tool for uncovering the principles of neural information processing in both artificial and biological systems.

We provide an open-source and optimized Python package at \href{https://github.com/LouisJalouzot/MLEM}{https://github.com/LouisJalouzot/MLEM}.
How to install the package and examples of code to use it can be found in the repository.


%% file: methods.tex
\section{Methods and Materials}
\label{sec:methods}

We formalize multivariate encoding as a metric learning problem.
The Metric Learning Encoding Model (MLEM) approach, summarized in \cref{fig:mlem}, consists of four main steps:
(A) For a set of stimuli, pairwise distances are computed for different theoretical features, creating a Representational Dissimilarity Matrix (RDM) $D^f$ for each.
(B) An RDM $D^N$ is also computed from neural representations of the same stimuli.
(C) A metric over the feature space is learned that maximizes the Spearman correlation between feature-based distances and neural distances. This metric is parameterized by a symmetric positive definite matrix $W$, whose off-diagonal terms explicitly model feature interactions.
(D) Permutation Feature Importance is used to assess the contribution of each feature and interaction to the model's predictions.
The following sections detail each of these steps, as well as the training procedure, the baseline and datasets used in this study.

\subsection{Metric Learning Encoding Models (MLEM)}

Formally, consider $n$ stimuli $(s_i)_{i\leq n}$ (e.g., sentences, images) for which values are available for each of $m$ theoretical features $(f_k)_{k\leq m}$ (e.g., linguistic features, visual categories, confounds).
The features can be continuous, discrete, categorical, ordered or not, and even multi-dimensional.
Assume that high-dimensional (neural) representations $Y\in \mathbb{R}^{n \times d}$ (e.g., fMRI images, LLM embeddings) are available for each stimulus $s_i$.

\subsubsection{Representational Dissimilarity Matrices (RDMs)}

Similarly to RSA, MLEM relies on computing RDMs (pairwise distances) between stimuli.
On one hand, in the neural space for a given distance (e.g. Euclidean):
$$D_{ij}^N = \left\|Y_i - Y_j\right\|_2$$
Where $Y_i$ is the neural representation of stimulus $s_i$ (e.g. fMRI image, LLM embedding).
The choice of Euclidean distance is standard in many representation analysis studies.
It assumes that neural representations form a vector space where proximity reflects similarity.
While other distance metrics (e.g., cosine distance) could be used, Euclidean distance is a general-purpose choice that is sensitive to both the angle and magnitude of representation vectors.

On the other hand, feature distances for all pairs of stimuli can be computed.
This work mainly explored nominal features (unordered and categorical) and therefore defined the distance between two stimuli to 1 if they have different values with respect to a given feature and 0 otherwise.
For ordered features, the absolute difference between the values of the two stimuli is used:

\begin{equation*}
    D_{ij}^{k} =
    \begin{cases}
        \left|f_k(s_i) - f_k(s_j)\right| & \text{if } f_k \text{ is ordered}\\
        \\
        \mathbb{1}\left[f_k(s_i) \neq f_k(s_j)\right] & \text{otherwise}
    \end{cases}
\end{equation*}

This choice is arbitrary and other distances could be used depending on the nature of the features.
For instance, multidimensional features could be used along with the Euclidean distance.

Note that some features might be underspecified, or undetermined, for some stimuli (e.g. gender value in the case of plural pronouns in English; `we', `they'..).
Here, in practice, this is implemented with NaN values and the distance is set to 0 if either stimulus of a pair is NaN. Here too, other choices can be made depending on one's dataset and goals.

Finally, let $D_{ij}^F$ be the $m$-dimensional vector of feature distances for a pair of stimuli $(s_i,s_j)$:
$$D_{ij}^F=\left( \ D_{ij}^{f_1} \ , \ D_{ij}^{f_2} \ , \ \ldots \ , \ D_{ij}^{f_m} \ \right)$$
It summarizes the theoretical distance between a pair of stimuli with respect to all features.

\subsubsection{Metric learning formulation}

The main goal of MLEMs is to learn a metric (distance) function, which approximates the neural distances from the theoretical distances.
The metric function is defined over all features, and it is parametrized using a weight matrix $W$.

We define MLEM as a metric function implemented
via a weighted norm $\|\bullet\|_W$.
That is, given a vector $D_{ij}^F$ for the theoretical distance between a pair of stimuli, we define the following metric function:

$$\widehat{D_{ij}^N}=\|D_{ij}^F\|_W = \sqrt{(D_{ij}^F)^T W (D_{ij}^F)}$$

The norm in this metric is parametrized by a matrix $W$, which weighs each feature and each feature-feature interaction.
We optimize $W$ to maximize the Spearman correlation $\rho$ between the empirical neural distances $D_{ij}^N$ and the modeled ones $\widehat{D_{ij}^N}$:

$$\sup_{W\in\mathbb{S}^{++}_m} \rho_{i<j}\left( \ \widehat{D_{ij}^N} \ , \ D_{ij}^N \ \right)$$

Note that the weight matrix needs to be symmetric positive definite $W \in\mathbb{S}^{++}_m$ for the norm to be valid (i.e. to satisfy the properties of a metric, such as positive definiteness and the triangle inequality).
This Metric Learning component is the core of MLEM.

The choice of Spearman correlation as the optimization objective is motivated by its robustness.
As a rank-based measure, it is invariant to monotonic transformations of the distances and less sensitive to outliers than other objectives like Pearson correlation or Mean Squared Error (MSE).
Pearson correlation assumes a linear relationship between the predicted and observed distances, which may be too restrictive.
MSE would aim for a direct match in distance values, making the model highly sensitive to the scale of the distances and to outliers.
Furthermore, Spearman correlation provides a score between -1 and 1, which is more directly interpretable than unbounded metrics like MSE or R$^2$ (not lower-bounded).
While Kendall's Tau is another robust rank-based correlation, Spearman correlation is often preferred for its computational efficiency on larger datasets.
By optimizing for rank correlation, MLEM focuses on preserving the relative arrangement of stimuli in the representation space, which is a core principle of second-order isomorphism methods.

\subsection{Model Training}

The model is trained using stochastic gradient descent.
Further details on the optimization procedure are provided in \cref{sec:optimization}.

\subsubsection{Batch size and breaking free from quadratic complexity}

\label{sec:subsampling}

The major limitation of methods relying on pairwise comparisons between stimuli, such as computing RDMs, is the quadratic complexity.
However here, the matrix of weights $W$ is optimized through stochastic gradient descent on batches of pairs of stimuli.
A training step only requires a subset of the RDMs corresponding to a batch of pairs.
This can be computed on the fly from the values of the features $f_k(s_i)$ and the high-dimensional representations $Y$.
Empirically, the optimization of the weight matrix $W$ often converges before the entire RDM has been explored.
This removes the need for the computation of all pairwise distances which can be prohibitive for large datasets.

One question remains: how to choose the batch size?
Such a parameter should be small for computing efficiency but large enough to provide a good coverage of the values of the theoretical features, in particular the sparse ones and their interactions.
In practice we used the smallest batch size that would allow for computing correlations between the features with variability below a given threshold.
This criterion ensures that the gradient estimates during optimization are based on a representative sample of feature co-occurrences, which is necessary for stable learning of the interaction terms in $W$.
More details about this procedure can be found in \cref{sec:batch_size_selection}.

\subsection{Evaluation and Permutation Feature Importance}

Models are evaluated using Spearman correlation $\rho$ (higher is better) between the predicted neural distances $\widehat{D_{ij}^N}$ and the empirical ones $D_{ij}^N$ on pairs of held-out stimuli.

Furthermore, we want to quantify the importance of each feature for the model, defined as how much each feature helps approximate neural distances.
To this end, {\em Permutation Feature Importance} \citep{breiman_random_2001} is used.
It measures the importance of a feature as the average degradation in model performance when permuting the values of the feature across samples.
This method is more reliable than looking directly at the weights of the model and provides quantities which are straightforward to interpret.
For instance, in this setting, an importance of 0.3 for one feature means that the model loses 0.3 of Spearman correlation when essentially destroying this feature.
This provides a clear and interpretable measure of importance, unlike metrics such as R² or MSE, for which the scale of the importance score is less intuitive.

More formally, assume a set of $n$ input samples $x_i\in\mathbb{R}^d$, with corresponding targets $y_i\in\mathbb{R}^{d'}$, a (trained) model $M:\mathbb{R}^d \to \mathbb{R}^{d'}$ that predicts targets from inputs, and a score function $S:\mathbb{R}^{n\times d'}\times\mathbb{R}^{n\times d'} \to \mathbb{R}$ that evaluates the model performance.
Permutation Feature Importance breaks the relationship between an input feature and the target by permuting the feature values across all samples.
If $S_\text{baseline}$ is the score of the model without any permutation:
$$S_\text{baseline} = S\Bigl(M\Bigl(X\Bigr),Y\Bigr)$$
and $S_{p}^k$ is the score of the model after applying permutation $p$ to the values of feature $f_k$ across all samples:
$$S_{p}^k = S\left(M\left(X^{f_k}_{p}\right),Y\right)$$
then the importance of feature $f_k$ is quantified as:
$$\mathbb{E}_{p}\left[S_\text{baseline} - S_{p}^k\right]$$

In practice, we find low variability across permutations therefore we approximate the expectation over $p$ by averaging over only 10 random permutations for computational efficiency.
The Permutation Feature Importance framework is extended to quantify the importance of feature interactions, which correspond to the off-diagonal terms in the weight matrix $W$.
This involves reshaping the input space to treat each interaction as a separate feature.
A detailed explanation of this procedure is provided in \cref{sec:pfi_details}.

Ablation studies (leave-one-covariate-out, \cite{lei_distributionfree_2018}) are another approach to feature importance which are more computationally expensive since they require retraining the model.
They are also unsuitable here since ablating/removing a single feature would require removing its interactions as well to be able to retrain with the parametrization on $W$.
Therefore it would be impossible to isolate the importance of a single feature or an interaction.

However Permutation Feature Importance is prone to false detections in the presence of high correlations between the features \citep{strobl_conditional_2008}.
Other methods are more robust to this, such as feature grouping \citep{chamma_variable_2024} and Conditional Permutation Importance (CPI, \cite{strobl_conditional_2008}), but they are not applicable here.
In the NLP examples, grouping features would lead to groups that are hard to interpret.
Regarding CPI, the gist of this method is to permute only the {\em independent} part of the features with respect to the other ones.
To do so, the method relies on training a predictor to reconstruct a feature based on the others and then permuting the residuals (intrinsic information from the feature).
In the MLEM framework, since the objective is to model feature interactions, the residuals would be null.
Indeed, predicting the interactions between $f_1$ and $f_2$ given at least $f_1$ and $f_2$ is very easy for a general predictor.

\subsection{Baseline: FR-RSA with interactions (FR-RSA-I)}

By design MLEM has parameters for each feature interaction (off-diagonal elements of $W$) and Permutation Feature Importance allows their importance to be quantified.
It can be critical to account for feature interactions in domains such as natural language processing.
For a fair comparison, an augmented variant of FR-RSA that includes interaction terms between features is considered as a state-of-the-art baseline.
We denote it as \textbf{FR-RSA-I}.
In practice, FR-RSA-I is the same as MLEM without the symmetric positive definite (SPD) constraint on $W$.
It is hypothesized that enforcing this constraint in MLEM is beneficial for a more reliable estimation of $W$.
The model is constrained to be a distance, which should help predict the target distance in neural space more precisely and more sample efficiently.
It is also hypothesized that it makes the model more robust to noise and overfitting.

\subsection{Datasets}

MLEM is evaluated in two setups, using two types of datasets: (1) synthetic datasets and (2) a fully annotated dataset of English sentences for which representations are extracted from a Large Language Model (LLM).
The reason to evaluate MLEM on synthetic datasets is that the true effects of features, and their interactions, on neural distances are known.
That is, all values of the elements of $W$ are known, to which we refer as the \textit{ground-truths}.
This allows assessment of MLEM's ability to recover a known underlying metric.
The sentence dataset, on the other hand, provides a more realistic setting to test the method on empirical data.

\subsubsection{Simulated datasets}

To evaluate MLEM's ability to recover a ground-truth metric over features and their interactions, synthetic datasets are generated with a known ground-truth weight matrix $W_{GT}$.
We consider $n=256$ stimuli ($s_i$) and $m=16$ binary theoretical features ($f_k$).
The features have two categories, ``A'' and ``B", which are randomly assigned to each stimulus, such that $f_k(s_i)$ takes the value ``A'' with a probability of 0.5 and ``B'' otherwise.
From these features, feature RDMs $D_{ij}^{k}$ are derived using an indicator distance (1 if feature values differ, 0 otherwise).

To establish a true underlying metric in the feature space, a ground truth symmetric positive definite matrix $W_{GT} \in \mathbb{S}^{++}_m$ is generated using \verb+make_spd_matrix+ from \verb+scikit-learn+ \citep{pedregosa_scikitlearn_2011} and normalized to have a Frobenius norm of 1 for comparability across datasets.
To generate neural representations that reflect this feature space, pairwise neural distances $D_{ij}^{N_{GT}}$ between stimuli $s_i$ and $s_j$ are computed as:
$$D_{ij}^{N_{GT}} = \|D_{ij}^F\|_{W_{GT}} = \sqrt{(D_{ij}^F)^T W_{GT} D_{ij}^F}$$
where $D_{ij}^F = [D_{ij}^f]_f$ is the vector of feature distances for the pair of stimuli $(s_i,s_j)$.
To create high-dimensional representations, Multidimensional Scaling (MDS) is used to generate $d=768$-dimensional representations $Y$ whose pairwise distances closely match $D^{N_{GT}}$.
To simulate realistic neural variability, Gaussian noise is added to these representations.
This noise is scaled by the standard deviation of each dimension of the generated representations and an additional configurable noise level parameter called "Noise level" in the figures.
Finally, to obtain the actual neural distances $D_{ij}^N$, the pairwise Euclidean distances between these noisy, high-dimensional representations are computed.

Models are trained on five datasets generated with different random seeds, using an 80\% training and 20\% testing split.
Metrics are reported as their average (full line) and their standard deviation (shaded area) across the five seeds.
Figures display the metrics measured on the test set when applicable.

\subsubsection{Relative Clause Sentence Dataset}

To test and compare MLEM and FR-RSA-I on real data, we use a dataset of generated sentences emphasizing a specific linguistic structure: relative clauses.
Following syntactic theory, relative clauses create sentences that are relatively complex, in particular object relative clauses due to a syntactic movement.
We focus are interested in relative clauses since this movement is expected to generate large distances in neural space.
We therefore contrast sentence stimuli with and without such syntactic movement.
In addition, we mark all sentence stimuli also based on other, presumably simpler, features, such as grammatical number (singular/plural).

A dataset of 7.7k sentences was generated with manually crafted grammars to emphasize the relative clause linguistic structure in a 2x2 design: center-embedded vs peripheral (right-branching) and subject vs object relative clauses.
This design is summarized in \cref{tab:rc_design}.
The dataset has 12 features which are displayed in \cref{tab:rc_features} along with their values, the distance function used, and a brief description.
Feature RDMs are then computed using either the indicator function (categorical values) or the absolute difference (continuous values) depending on the feature.

\begin{table}[!ht]
    \centering
    \renewcommand{\arraystretch}{2}
    \begin{tabular}{ccrl}
        \toprule
        \multirow{1}{*}{\textbf{\quad Attachment Site \qquad}}
        & \multirow{1}{*}{\textbf{Relative Clause Type}} & \\
        \toprule
        \multirow{4}{*}{\textbf{Center-Embedded}}
        & \multirow{2}{*}{\textbf{Subject relative}}
        & \multirow{1}{*}{Template:}
        & \multirow{1}{*}{\textit{Subj \textcolor{purple}{[who verb obj]} verb obj.}} \\
        & & \multirow{1}{*}{Example:}
        & \multirow{1}{*}{\textcolor{gray}{The woman \textcolor{purple}{[who sees the princess]} admires the actress.}} \\
        \cline{2-4}
        & \multirow{2}{*}{\textbf{Object relative}}
        & \multirow{1}{*}{Template:}
        & \multirow{1}{*}{\textit{Subj \textcolor{purple}{[who subj verb]} verb obj.}} \\
        & & \multirow{1}{*}{Example:}
        & \multirow{1}{*}{\textcolor{gray}{The woman \textcolor{purple}{[who the princess sees]} admires the actress.}} \\
        \midrule
        \multirow{4}{*}{\makecell[t]{\textbf{Peripheral} \\ \textbf{(Right-Branching)}}}
        & \multirow{2}{*}{\textbf{Subject relative}}
        & \multirow{1}{*}{Template:}
        & \multirow{1}{*}{\textit{Subj verb obj \textcolor{purple}{[who verb obj]}.}} \\
        & & \multirow{1}{*}{Example:}
        & \multirow{1}{*}{\textcolor{gray}{The woman sees the princess \textcolor{purple}{[who admires the actress]}.}} \\
        \cline{2-4}
        & \multirow{2}{*}{\textbf{Object relative}}
        & \multirow{1}{*}{Template:}
        & \multirow{1}{*}{\textit{Subj verb obj \textcolor{purple}{[who subj verb]}.}} \\
        & & \multirow{1}{*}{Example:}
        & \multirow{1}{*}{\textcolor{gray}{The woman admires the actress \textcolor{purple}{[who the princess sees]}.}} \\
        \toprule
    \end{tabular}
    \vspace{0.25cm}
    \caption{2x2 design of the relative clause sentence dataset to emphasize center-embedded vs peripheral and subject vs object relative clauses.}
    \label{tab:rc_design}
\end{table}

Permutation Feature Importance is known to be sensitive to high correlations between features.
Because importances are harder to interpret under such correlations, the dataset was carefully balanced to mitigate them.
As a result, the correlations between the feature RDMs are negligible (c.f. \cref{fig:rc_correlations}).

For this sentence dataset, representations are taken from a well-studied Large Language Model: BERT \citep{devlin_bert_2019}.
More precisely, the pre-trained \verb+bert-base-uncased+ model from the \verb+transformers+ library \citep{wolf_transformers_2020}.
BERT is fed with the sentences and hidden states are extracted from its 12 layers.
These hidden states are averaged over the token dimension to get fixed-size sentence embeddings, which is necessary for the method.
These embeddings are then regarded as the neural representations of the sentences for BERT.
Neural distances $D_{ij}^N$ are computed as the Euclidean distance between these sentence embeddings.
To also evaluate robustness to noise in this more realistic setting, we added artificial Gaussian noise to the BERT representations in a similar way as the simulation.
The noise is scaled by the standard deviation of the representations for each dimension and a configurable noise level parameter, called "Artificial Noise level" in the figures.

5-fold cross-validation is applied and metrics are reported as their average (full line) and their standard deviation (shaded area) across the five folds.
Figures display the metrics measured on the test set when applicable.

In practice we run the batch size estimation procedure described in \cref{sec:subsampling} and \cref{sec:optimization} with a conservative set of parameters and obtain a large batch size of $\sim$20,000 for this dataset for high accuracy.

%% file: results.tex
\section{Results}
\label{sec:results}

\subsection{MLEMs are more accurate and robust to noise compared to the strongest baseline method}

A key difference between MLEM and FR-RSA-I is that MLEM enforces the weight matrix $W$ to be symmetric positive definite (SPD), constraining it to define a valid metric.
We hypothesize that this constraint provides regularization that leads to more accurate weight recovery and greater robustness to noise, as demonstrated in the following results.

\begin{figure}[!th]
    \centering
    \begin{subfigure}[b]{0.49\textwidth}
        \includegraphics[width=\textwidth]{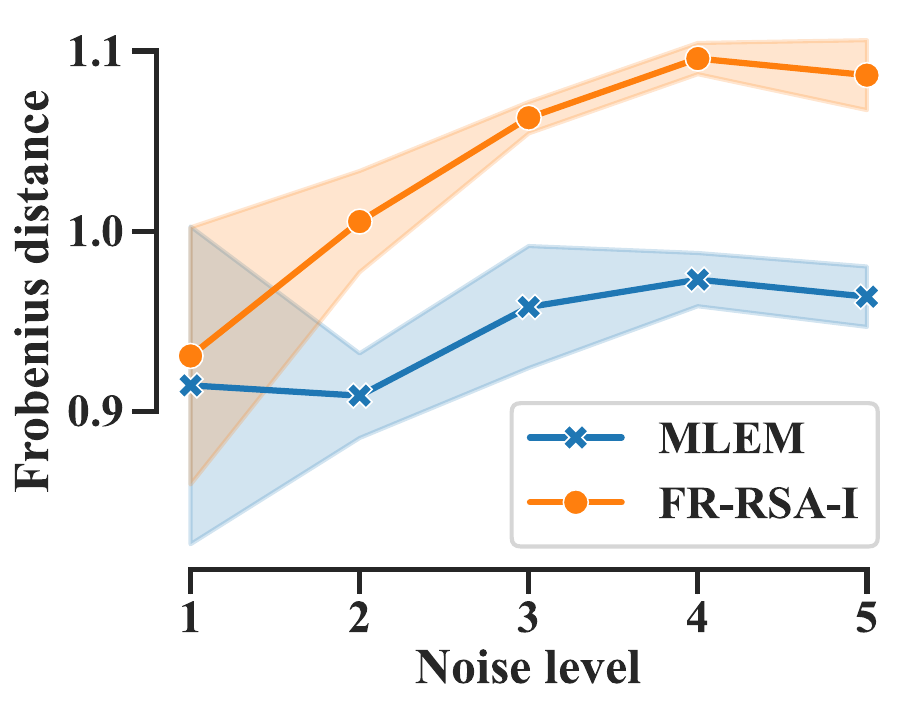}
        \caption{Accuracy on simulated data.}
        \label{fig:fro_ground_truth}
    \end{subfigure}
    \hfill
    \begin{subfigure}[b]{0.49\textwidth}
        \includegraphics[width=\textwidth,trim=0 0 0 1.95cm,clip]{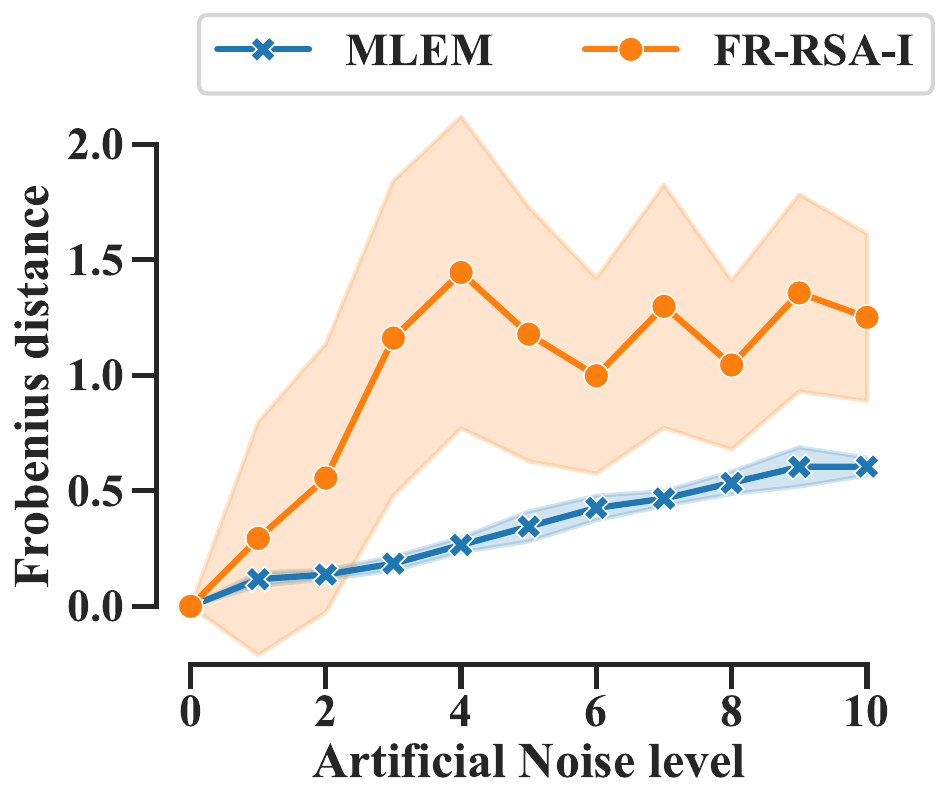}
        \caption{Robustness to artificial noise.}
        \label{fig:fro_noiseless}
    \end{subfigure}
    \caption{
        \textbf{MLEM is more accurate and robust to noise than FR-RSA-I.}
        Frobenius distance at different noise levels between estimated weights and \textbf{(a)} ground-truth weights in synthetic data (accuracy), and \textbf{(b)} estimation at noise level 0 on LLM embeddings (robustness).
        MLEM (blue) consistently achieves lower error than FR-RSA-I (orange).
        In both plots, the full line represents the average and the shaded area represents the standard deviation across 5 runs.
    }
    \label{fig:fro_gt_and_noise}
\end{figure}

\paragraph{Simulation Results.}
First, we used simulated datasets where the ground-truth weight matrix $W_{GT}$ is known (see \cref{sec:methods}).
\cref{fig:fro_gt_and_noise}a shows the Frobenius distance between the estimated weights and the ground-truth weights for both methods across different levels of noise.
MLEM consistently achieves a lower Frobenius distance, indicating that its estimated weights are closer to the ground truth and demonstrating its superior accuracy in recovering the true feature geometry.

\paragraph{Experiments with LLMs.}
Second, to evaluate robustness, we measured how much the estimated weights changed when artificial noise was added to the neural representations from layer 7 of the LLM.
\cref{fig:fro_gt_and_noise}b shows the Frobenius distance between weights estimated with and without this noise.
The weights from MLEM show a smaller distance to their noiseless counterpart across all noise levels, demonstrating the superior stability and robustness of MLEM.

To provide a more detailed view of this robustness, we examined the profiles of learned weights and feature importances under increasing levels of artificial noise (\cref{fig:weights_noise,fig:fi_noise}).
For both weights and feature importances, the profiles for FR-RSA-I degrade more rapidly than for MLEM, confirming MLEM's superior stability.
To quantify the stability of the feature rankings, we used Kendall's weighted $\tau$ to compare feature importance profiles at each noise level to the noiseless profile (\cref{fig:weighted_tau_noiseless}).
MLEM maintains a consistently higher correlation, indicating that its feature importance rankings are more reliable in noisy conditions.

\begin{figure}[!th]
    \centering
    \begin{subfigure}[b]{0.49\textwidth}
        \includegraphics[width=\textwidth]{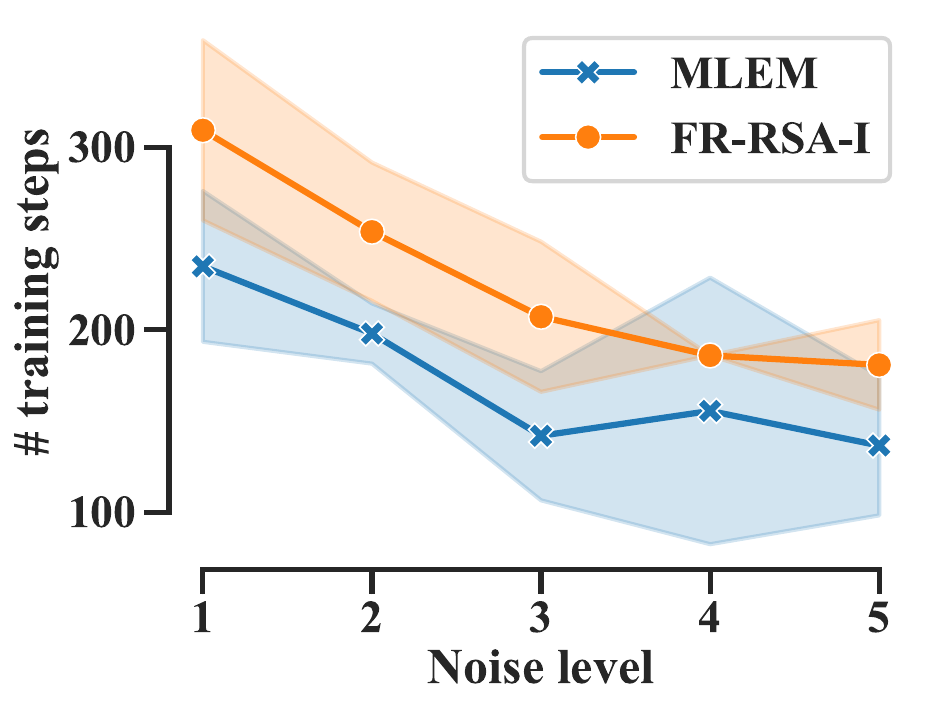}
        \caption{Simulated data.}
        \label{fig:n_epochs_simulation}
    \end{subfigure}
    \hfill
    \begin{subfigure}[b]{0.49\textwidth}
        \includegraphics[width=\textwidth,trim=0 0 0 1.65cm,clip]{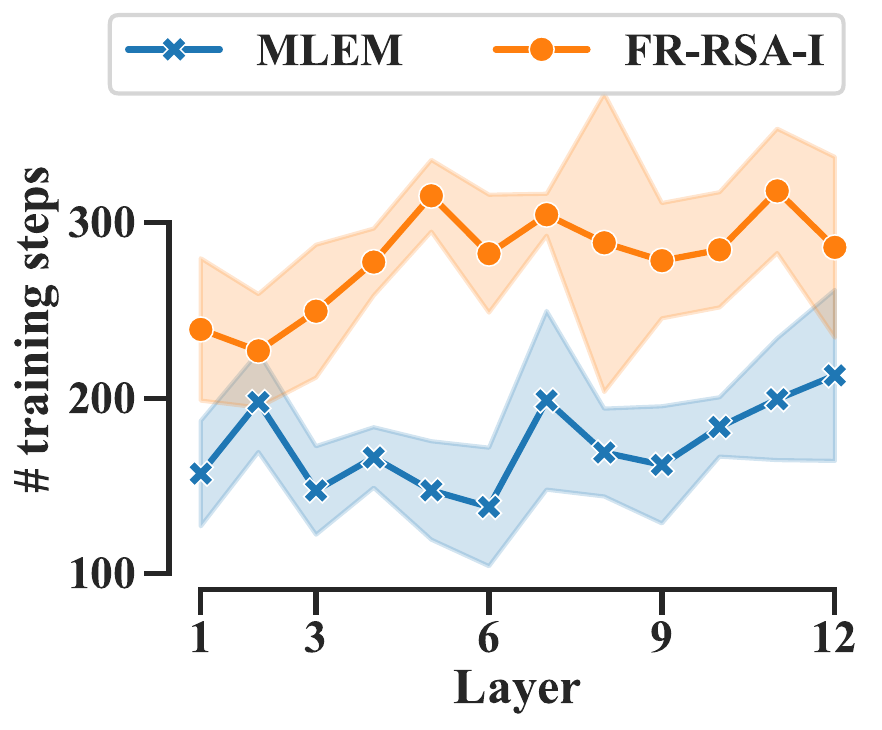}
        \caption{Relative clause dataset with LLM embeddings.}
        \label{fig:n_epochs_bert}
    \end{subfigure}
    \caption{
        \textbf{MLEM converges faster than FR-RSA-I.}
        The plots compare the number of training steps (epochs) of MLEM (blue) and FR-RSA-I (orange).
        \textbf{(a)} On simulated datasets, MLEM converges faster across all noise levels.
        \textbf{(b)} On the relative clause dataset with LLM embeddings, MLEM converges faster for each of the 12 layers.
        This demonstrates MLEM's superior computational efficiency.
        The full line represents the average and the shaded area represents the standard deviation across 5 runs.
    }
    \label{fig:n_epochs}
\end{figure}

\subsection{MLEMs converge faster than baseline methods}
We next compared the convergence speed of MLEMs and FR-RSA-I.
\cref{fig:n_epochs} shows the number of training steps for both methods on the simulated datasets (\cref{fig:n_epochs_simulation}) and the relative clause dataset with LLM embeddings (\cref{fig:n_epochs_bert}).
In both cases, MLEM converges faster than FR-RSA-I.
As our approach does not explore the whole RDMs before converging (see \cref{sec:subsampling}), this result also means that MLEM is more data efficient.

As shown in \cref{fig:n_epochs_noise}, MLEM's speed advantage is most pronounced at low artificial noise levels but both methods seem to converge at the same speed when noise increases.
However MLEM still shows less variability in the number of training steps across runs.


\subsection{Encoding performance is comparable between MLEM and FR-RSA-I}

MLEM and FR-RSA-I obtain similar final encoding performance measured by Spearman correlation between predicted and observed neural distances on held-out stimuli (see \cref{fig:spearman_all}).
Therefore MLEM recovers more faithful feature and interactions weights, yields more reliable feature importance profiles, and converges faster, while preserving encoding power.

\subsection{MLEMs reveal importance profiles of the theoretical feature across layers of an LLM}

To understand how the LLM represents sentences, we train one MLEM per layer and investigate the importance of features and interactions obtained.

\begin{figure}[!th]
    \centering
    \includegraphics[width=\textwidth]{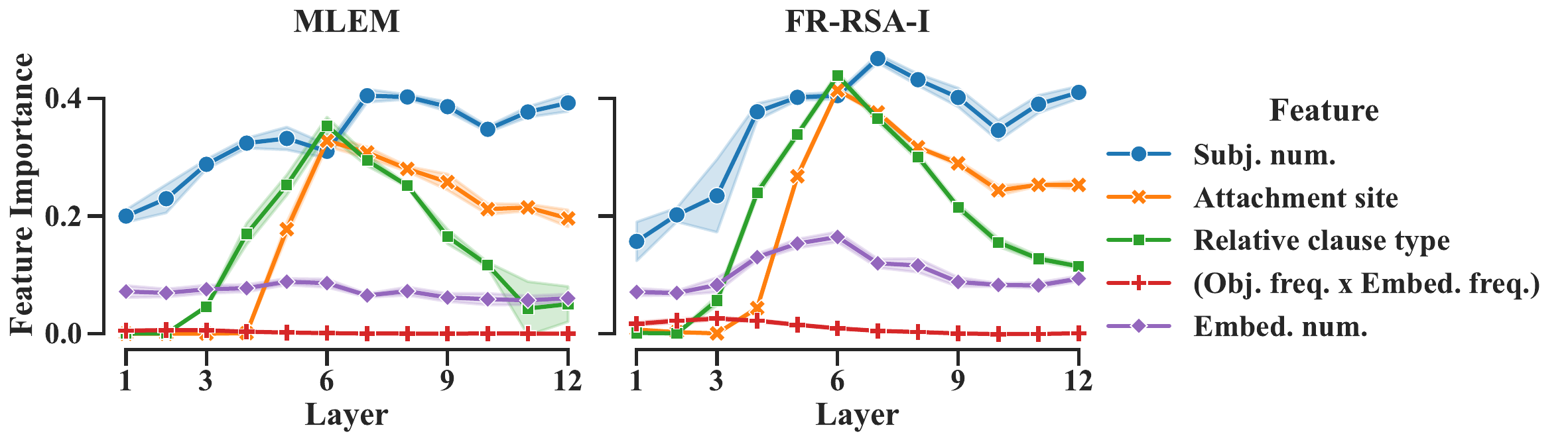}
    \caption{
        \textbf{MLEM and FR-RSA-I identify similar profiles of feature importance across LLM layers.}
        The plots show the top feature importances for MLEM (left) and FR-RSA-I (right) across the 12 layers of the LLM.
        Each line represents a feature or interaction, and its y-value indicates its importance at a given layer.
        Both methods reveal a similar pattern: features related to sentence structure, such as "Relative Clause type" and "Attachment site", gain importance in the middle layers, which is consistent with prior work on LLM's syntactic processing.
        This suggests that both methods can uncover meaningful linguistic information.
        The full line represents the average and the shaded area represents the standard deviation across 5 cross-validation folds.
    }
    \label{fig:feature_importance}
\end{figure}

For readability, we display a subset of the 12 features from the relative clause dataset.
We selected the top features in terms of both highest importance and absolute weight across layers for MLEM and FR-RSA-I, resulting in 4 features and one interaction.
\cref{fig:feature_importance} shows the profiles of feature importances across layers and \cref{fig:weights} shows the absolute weights.
The importance curves show interesting patterns.
For instance "Relative Clause type" and "Attachment site" achieve their highest scores in the middle layers, consistent with prior work suggesting that LLMs process syntax predominantly in middle layers (in particular \cite{hewitt_structural_2019} for BERT).

\subsection{Feature Importance of MLEMs reveal complex geometrical structures in LLM representations}

\begin{figure}
    \centering
    \includegraphics[width=\textwidth]{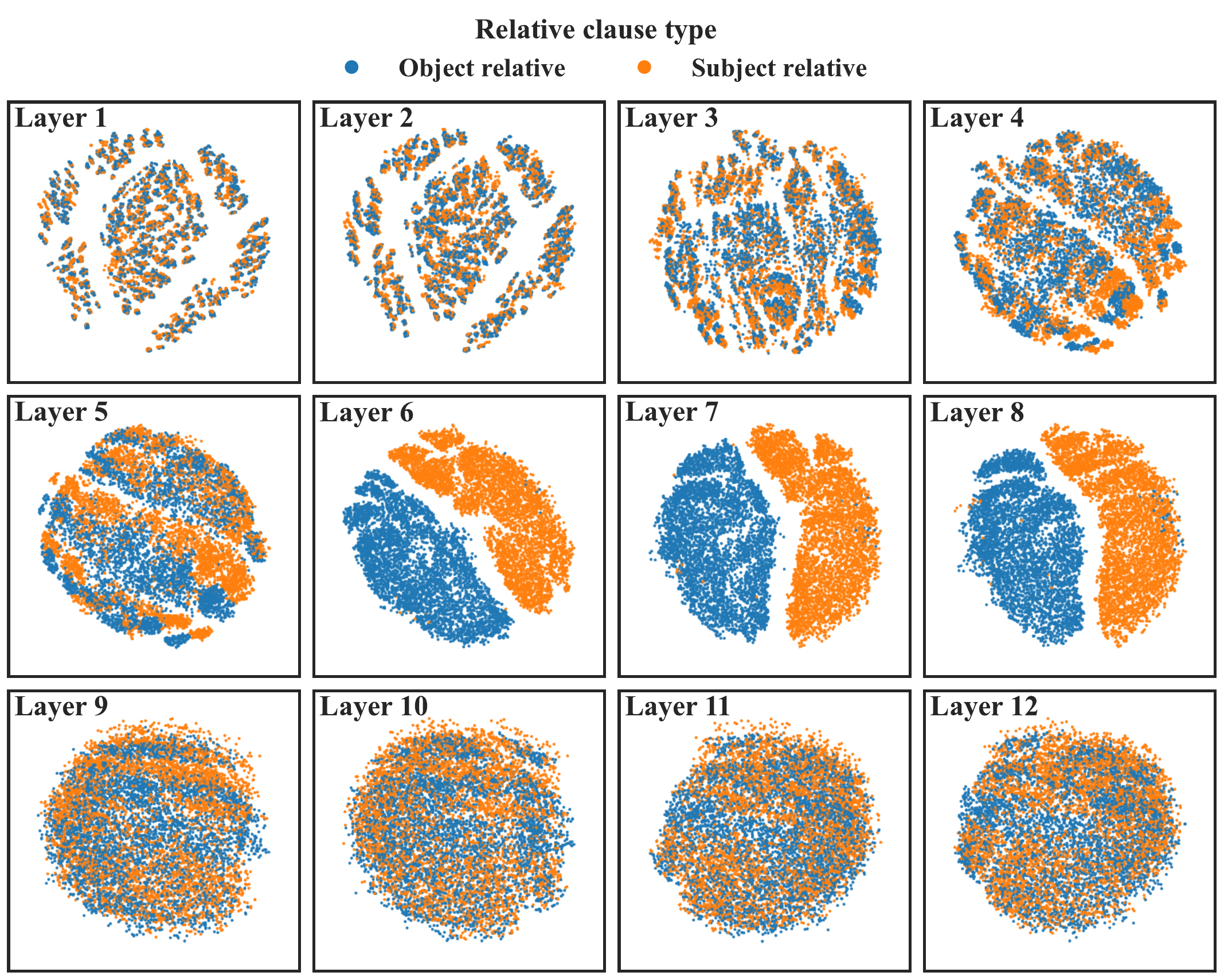}
    \caption{
        \textbf{The geometry of LLM representations reflects syntactic structure, especially in middle layers.}
        This figure shows 2D MDS visualizations of sentence representations from each of the LLM's 12 layers.
        Points are colored by the "Relative Clause type" feature (object relative: blue, subject relative: orange).
        A clear clustering based on this feature emerges in the middle layers, where the two colors become separable.
        This geometric separation aligns with the peak importance of the "Relative Clause type" feature shown in \cref{fig:feature_importance}, validating that the feature importances identified by the approach correspond to tangible structures in neural space.
    }
    \label{fig:mds_layers}
\end{figure}

To verify that importance values correspond to meaningful geometric structures, we visualized the representations in 2D using Multidimensional Scaling (MDS, \cite{kruskal_multidimensional_1964}).
The layer-wise feature importance profiles reflect the geometry of the neural representation.
\cref{fig:mds_layers} shows that a clear clustering based on "Relative Clause type" emerges in the middle layers, where its feature importance peaks in \cref{fig:feature_importance}.
This validates that feature importance profiles produced by MLEM are informative about the geometry of neural representations.

\begin{figure}
    \centering
    \includegraphics[width=\textwidth]{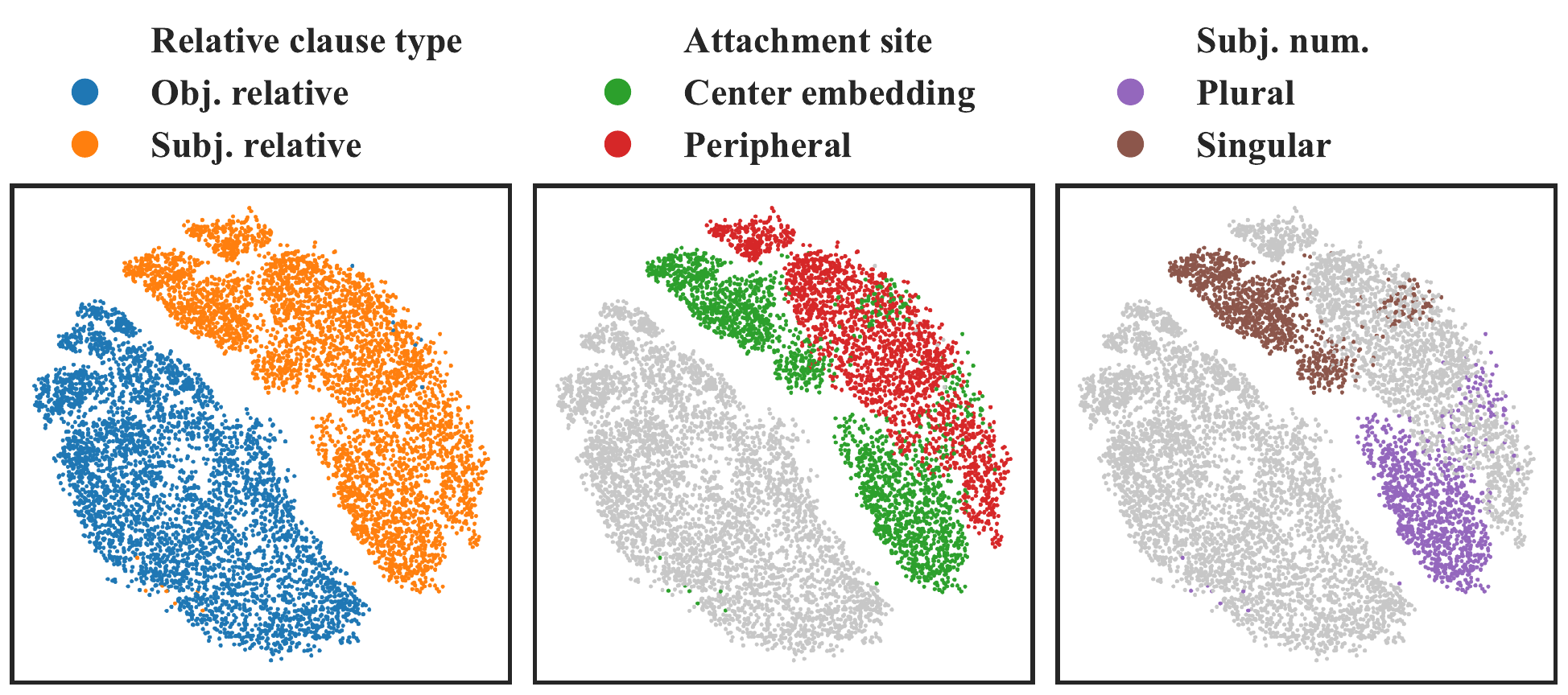}
    \caption{
        \textbf{Top features from MLEM reveal a hierarchical clustering in the space of neural representation.}
        This figure visualizes the sentence representations from layer 6 of the LLM, demonstrating how the top three features identified by MLEM for this layer organize the space hierarchically.
        First, representations cluster by "Relative Clause type" (subject relative: blue, object relative: orange).
        Within the subject-relative cluster, a secondary split emerges based on "Attachment site" (center-embedded: green, peripheral: red).
        Finally, within the subject-relative, center-embedded cluster, a third level of organization appears based on "Subject number" (plural: purple, singular: brown).
        This hierarchical structure, discovered using MLEM's feature importances, provides a clear and interpretable view of the geometry of neural representations.
        See \cref{fig:hierarchical_mds_full} for a complete view.
    }
    \label{fig:hierarchical_mds}
\end{figure}

Furthermore, the relative feature importances of different features reflect a nested-structure geometry.
\cref{fig:hierarchical_mds} demonstrates at the layer 6, the most important features define a hierarchical clustering.
The space is first organized by "Relative Clause type," then by "Attachment site," and finally by "Subject number," offering an interpretable visualization of how abstract linguistic properties are hierarchically encoded.

\cref{fig:mlem_frrsa_layers} compares MLEM and FR-RSA-I across layers (weights and feature importance agreement), and \cref{fig:mlem_frrsa_noise} extends this comparison under artificial noise.
Finally, \cref{fig:univariate} contrasts the multivariate MLEM with unit-wise univariate MLEMs, showing the benefit of modeling distributed representations.

To contextualize MLEM's informativeness against alternative approaches, we compared it with both decoding and multivariate encoding baselines.
A standard decoding approach, using linear classifiers to predict features from representations, confirms that feature information is present but reveals little about how representational geometry evolves across layers, as accuracies remain uniformly high (\cref{fig:decoding}).
In contrast, a multivariate random forest encoding baseline recovers a feature importance profile similar to MLEM's but cannot model feature interactions by design and lacks the robustness conferred by MLEM's metric learning framework (\cref{sec:encoding_baseline}).
These comparisons underscore MLEM's unique advantage in providing detailed and robust insights into the structure of neural representations.

%% file: discussion.tex
\section{Discussion}

This work introduced Metric Learning Encoding Models (MLEM), a multivariate encoding framework that learns an optimal metric over a feature space to model the geometry of neural representations.
Experiments on simulated and real-world data demonstrate that MLEM offers several advantages over existing methods like FR-RSA: it is more accurate in recovering ground-truth feature weights (\cref{fig:fro_ground_truth}), more robust to noise (\cref{fig:fro_noiseless}), and converges faster (\cref{fig:n_epochs_simulation,fig:n_epochs_bert}) which makes it more data-efficient.
By combining this with Permutation Feature Importance, MLEM reliably quantifies the contribution of both individual features and their interactions to neural representations, revealing interpretable geometric structures in high-dimensional data (\cref{fig:mds_layers,fig:hierarchical_mds}).

\subsection{Advantages of multivariate encoding over alternative approaches}

MLEM's multivariate approach offers several key advantages over decoding methods and univariate encoding approaches.
Decoding models (or "probes"), while demonstrating that feature-related information is present, are susceptible to confounds where a feature may be decodable simply because it correlates with another, truly encoded feature \citep{hewitt_designing_2019, belinkov_probing_2022, kumar_probing_2022}.
Moreover, as shown in \cref{fig:decoding}, decoding accuracies remain uniformly high across layers and provide little insight into how representational geometry changes, highlighting a key limitation of this approach.

Univariate encoding approaches, which fit each neural unit independently, cannot model distributed representations encoded across populations of units, a critical characteristic of neural computation \citep{georgopoulos_neuronal_1986}.
The comparison with univariate MLEMs (\cref{fig:univariate}) demonstrates that the multivariate approach successfully captures geometric structure encoded across multiple units, which individual univariate models cannot access.
This is particularly important for understanding complex domains like language, where information is often distributedly encoded across neural populations rather than localized in individual units.

\subsection{Modeling feature interactions through metric learning}

A central contribution of MLEM is its principled approach to modeling feature interactions through metric learning.
Feature-Reweighted RSA (FR-RSA) can be seen as a special case of MLEM where the weight matrix is constrained to be diagonal, corresponding to independent feature contributions without interactions.
Even when FR-RSA is artificially extended to capture interactions (FR-RSA-I), it fails to match MLEM's performance because the set of weights is not constrained to form a valid metric.
While our analysis of LLM representations did not reveal strong, interpretable interactions, the ability to model them is crucial.
Even if interactions are not of primary interest, their inclusion can lead to more accurate estimates of the main effects of individual features, as demonstrated in our simulations.

The symmetric positive definite (SPD) constraint in MLEM ensures that the learned weights define a valid metric, which serves multiple purposes.
First, it provides theoretical grounding by guaranteeing that the weighted distance satisfies metric properties such as positive definiteness and the triangle inequality.
Second, this constraint functions as a form of regularization, guiding the optimization toward more stable and interpretable solutions.
As demonstrated in our results (\cref{fig:fro_gt_and_noise}), this constraint is crucial for achieving superior accuracy in recovering ground-truth weights and increased robustness to noise compared to FR-RSA-I.
Without this constraint, as in FR-RSA-I, the learned weights may not correspond to a meaningful geometric structure, making the model more prone to overfitting and less robust to noise.
The uncoupled estimation of interaction terms in FR-RSA-I can impair their accuracy, as shown in our simulations.
The superior performance of MLEM in recovering ground-truth weights and its increased robustness can be attributed to this principled metric learning formulation.

\subsection{Robustness to noise in neural data}

Neural data, whether from biological systems or artificial networks, inevitably contains noise from various sources including measurement error, biological variability, and computational precision.
The robustness analysis (\cref{fig:fro_noiseless,fig:weights_noise,fig:fi_noise}) demonstrates that MLEM maintains more stable weight estimates and feature importance profiles under increasing noise levels compared to FR-RSA-I.
This robustness stems from the SPD constraint, which regularizes the learning process and prevents the model from fitting to spurious noise patterns.
Additionally, the use of Spearman correlation as the optimization objective makes the approach less sensitive to outliers and non-linear relationships between feature and neural geometries.
This is because Spearman correlation operates on ranks, making it invariant to monotonic transformations of the distances and robust to extreme values.
This choice also makes the resulting scores and feature importances more interpretable than alternatives like MSE or R$^2$.
This robustness is crucial for real-world applications where neural measurements are inherently noisy.

\subsection{Computational efficiency and practical impact}

The computational efficiency of MLEM stems from multiple innovations that make it convenient for both large and small datasets.

First, the use of stochastic gradient descent for the optimization along with the on-the-fly computation of batches of data (\cref{sec:subsampling}) avoids the quadratic complexity of computing full RDMs.
For instance on the relative clause dataset, MLEM converges in less than 200 steps (\cref{fig:n_epochs_bert}).
With the estimated batch size of 20,000 pairs, this means it explored at most 4 million of the $\sim$30 million possible pairs, or less than 15\% of the full RDMs.
This makes MLEM scalable to datasets with thousands or even millions of stimuli for which computing and storing full RDMs would be computationally intractable with traditional RSA-based methods.
Furthermore, convergence without needing to see all stimulus pairs makes MLEM more data-efficient.
This means that for limited data scenarios, which is common in neuroimaging studies where acquisition is expensive and time-consuming, MLEM converges to a better solution with the same amount of data.

Second, as demonstrated in \cref{fig:n_epochs_simulation,fig:n_epochs_bert}, MLEM converges faster than FR-RSA-I.
The sole difference between the two methods is the SPD constraint, hence the speed advantage can be directly attributed to this principled parametrization.
By constraining the optimization to the manifold of valid metrics, the SPD parametrization guides gradient descent toward meaningful solutions more efficiently than the unconstrained approach of FR-RSA-I.
Therefore the SPD constraint provides an additional performance boost in addition to its regularization benefits.

\subsection{Comparing MLEM with similar approaches from the literature}

The closest variant of RSA to MLEM is \textit{Representational Similarity Learning} (RSL, \cite{oswal_representational_2016}).
RSL is a decoding approach: given human similarity judgments $S$ and neural measurements $X$, it seeks a sparse and symmetric weight matrix $W$ such that $XWX^T \approx S$.
In contrast, MLEM is an encoding approach, neural distances are predicted from feature distances.
Furthermore, its focus is on the neural encoding of multiple theoretical features while RSL was applied to decode a single matrix of human similarity judgments.

Another related method is Mixed RSA \citep{khaligh-razavi_fixed_2017}, which combines voxel-receptive-field (RF) modeling with RSA.
Mixed RSA first learns a linear mapping from model features to individual voxel responses on a training set, and then computes an RDM from the predicted voxel responses on a test set.
This predicted RDM is then compared to the empirical RDM.
While both MLEM and Mixed RSA aim to find a better alignment between feature-based and neural representations, they operate at different levels.
Mixed RSA is a hybrid approach that first builds a first-order encoding model (features to activations) and then performs a second-order comparison (RSA).
In contrast, MLEM is a pure second-order model that directly learns a metric on the feature space to match the geometry of the neural space, without modeling individual neural units.
This makes MLEM more directly focused on the representational geometry and less dependent on the quality of individual unit predictions.

A similar metric learning approach was also adopted more practically in the study of phoneme perception, based on behavioral phoneme-confusion data \citep{lakretz_metric_2018}.

Finally, other multivariate encoding approaches include Reduced-Rank Ridge Regression (RRR, \cite{mukherjee_reduced_2011}) and Back-to-back (B2B) regression \citep{king_backtoback_2020}.
Both are first-order encoding models that predict neural activations directly from features, unlike MLEM, which is a second-order model that operates on distances.
RRR regularizes the linear mapping from features to neural activations by enforcing a low-rank constraint on the coefficient matrix. This is conceptually different from MLEM's approach of learning a metric on the feature space.
B2B regression aims to disentangle correlated features by chaining a decoding model (from neural data to features) with an encoding model (from features back to the decoder's predictions). While it addresses feature collinearity, it does not explicitly model feature interactions or the geometry of the representation space in the way MLEM does.

\subsection{Broader impact and future applications}

The MLEM framework is modality-agnostic and opens several avenues for future research across multiple domains.
In neuroscience, applying MLEM to human brain data (fMRI, MEG, ECoG) could help bridge the gap between findings from computational models and the principles of neural computation in biological systems.
In artificial intelligence, MLEM provides a powerful tool for interpreting the internal representations of deep learning models across various modalities.

The method's ability to uncover hierarchical clustering based on feature importance (\cref{fig:hierarchical_mds}) demonstrates its potential for discovering interpretable structure in high-dimensional representations.
This capability could prove valuable for understanding how abstract concepts are organized in both artificial and biological neural systems.


\subsection{Limitations}

\paragraph{Large and Heterogeneous Datasets}
As explained in \cref{sec:subsampling}, the implementation of MLEM uses a subsampling strategy to avoid the quadratic complexity of pairwise comparisons.
Optimization often converges long before the entire set of stimulus pairs has been seen.
This efficiency, however, poses a risk for very large and heterogeneous datasets.
If convergence is reached too early, some informative stimulus pairs, especially those related to sparse features, may not be sampled, potentially biasing the learned metric.
While the adaptive batch sizing procedure (\cref{sec:subsampling}) is designed to mitigate this by ensuring feature correlations are stable within batches, it may lead to computationally expensive large batches for highly diverse datasets.

\paragraph{Weighting Units in the Neural Space}
One could consider extending MLEM to learn weights for individual units (e.g., neurons or voxels) in the neural space, similar to some variants of RSA \citep{oswal_representational_2016}.
For instance one could learn a diagonal matrix $\lambda$ to compute a weighted neural distance: $D_{ij}^N = \left\|Y_i - Y_j\right\|_\lambda$.
While this could improve the alignment between feature and neural spaces (i.e., achieve a higher Spearman correlation), it would come at the cost of interpretability.
The resulting feature importances would explain a distorted or degenerated neural space, making it difficult to draw conclusions about the original representations.
For spatial data like fMRI, a searchlight analysis remains a more advisable approach for localizing feature information.
If such a feature were to be implemented, regularization on $\lambda$ would be necessary to prevent it from becoming overly sparse, which would amount to selecting only a few neural units.

\section{Conclusion}

This work presented MLEM, a novel framework for interpreting high-dimensional representations by modeling their geometry as a learned metric over theoretical features.
By explicitly accounting for feature interactions through a principled metric learning approach and leveraging a robust optimization objective, MLEM provides more accurate, stable, and efficient interpretations than previous methods.
Its effectiveness has been demonstrated in both controlled simulations and in analyzing the representations of a large language model.
MLEM represents a significant advance over existing approaches by combining the strengths of encoding models with the ability to capture distributed neural representations and modeling feature interactions.
The framework's versatility makes it a powerful tool for researchers seeking to uncover the principles of neural information processing in both artificial and biological systems.

%% file: appendix.tex
\section{Relative Clause dataset}

\subsection{Dataset construction and balancing}

\begin{figure}[!th]
    \centering
    \includegraphics[width=.8\textwidth]{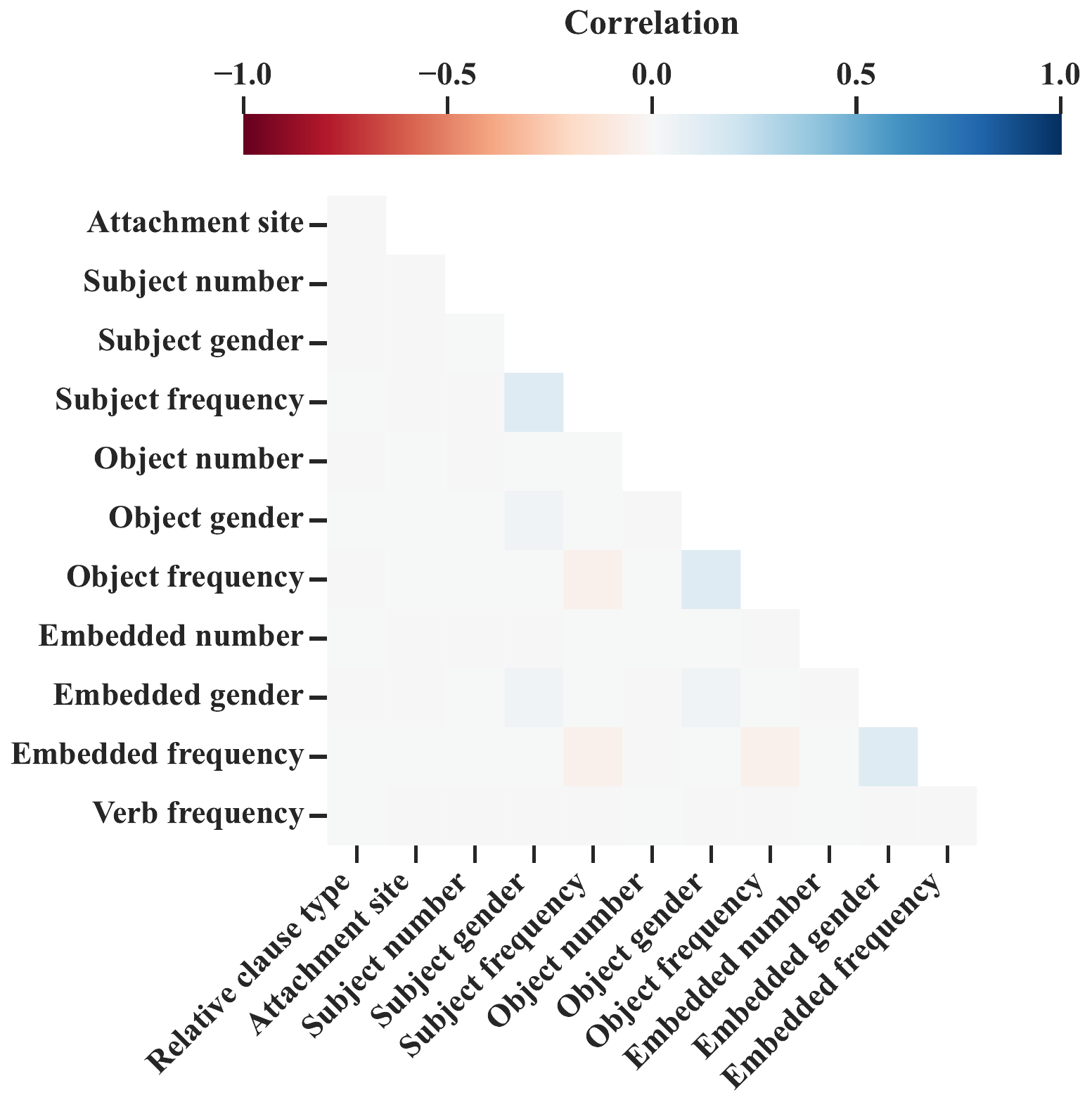}
    \caption{
        \textbf{Feature RDMs in the Relative Clause dataset are uncorrelated.}
        This heatmap shows the Pearson correlation between the Representational Dissimilarity Matrices (RDMs) of the 12 linguistic features in the dataset.
        The correlations are computed over all pairs of stimuli.
        The low values across the matrix (the largest being 0.125) indicate that the dataset is well-balanced and that the features are largely independent.
        This is crucial for the interpretability of methods like Permutation Feature Importance, as it minimizes the risk of attributing importance to confounded features.
    }
    \label{fig:rc_correlations}
\end{figure}

The Relative Clause dataset was generated to emphasize syntactic variation in a 2×2 design that manipulates embedding site (center-embedded vs peripheral) and relative-clause role (subject vs object).
Sentences were produced using controlled grammars to fill a combinatorial space of linguistic features; the final dataset contains approximately 7.7k sentences and 12 features (listed in \cref{tab:rc_features}).
Features include binary syntactic properties (e.g., relative-clause type, attachment site), morphological attributes (subject/object/embedded number and gender), lexical frequency bins (Zipf scores), and the verb lemma.
The dataset was balanced with respect to these features to avoid spurious correlations between feature RDMs; \cref{fig:rc_correlations} shows the (near-zero) correlations across feature RDMs, indicating a well-balanced design where permutation-based importances remain interpretable.

\begin{table}[h!]
    \centering
    \renewcommand{\arraystretch}{1.5}
    \begin{tabular}{lllp{7.5cm}}
        \toprule
        \textbf{Feature} & \textbf{Values} & \textbf{Distance} & \textbf{Description} \\
        \midrule
        Relative Clause type & Subject, Object & Indicator & Whether the relative clause modifies the subject or the object of the main clause.\\
        Attachment site & Peripheral, Center-embedded & Indicator & Position of the relative clause relative to the main clause.\\
        Subject number & Singular, Plural & Indicator & Grammatical number of the main clause subject.\\
        Subject gender & Feminine, Masculine & Indicator & Grammatical gender of the main clause subject.\\
        Subject frequency & 4.0, 4.5, 5.0, 5.5 & Euclidean & Zipf frequency of the main clause subject.\\
        Object number & Singular, Plural & Indicator & Grammatical number of the main clause object.\\
        Object gender & Feminine, Masculine & Indicator & Grammatical gender of the main clause object.\\
        Object frequency & 4.0, 4.5, 5.0, 5.5 & Euclidean & Zipf frequency of the main clause object.\\
        Embedded number & Singular, Plural & Indicator & Grammatical number of the embedded clause subject.\\
        Embedded gender & Feminine, Masculine & Indicator & Grammatical gender of the embedded clause subject.\\
        Embedded frequency & 4.0, 4.5, 5.0, 5.5 & Euclidean & Zipf frequency of the embedded clause subject.\\
        Verb lemma & admire, see & Indicator & Lemma of the verb.\\
        \bottomrule
    \end{tabular}
    \vspace{0.25cm}
    \caption{Linguistic features and their values in the Relative Clause dataset.}
    \label{tab:rc_features}
\end{table}

\clearpage
\section{Optimization Details}
\label{sec:optimization}

Optimizing the Spearman correlation while constraining the weight matrix $W$ to be symmetric positive definite (SPD) is a non-trivial problem.
This optimization is performed through stochastic gradient descent on a differentiable relaxation of the Spearman correlation introduced by \cite{blondel_fast_2020} and we used the optimized PyTorch implementation from \href{https://github.com/teddykoker/torchsort}{torchsort}.

To enforce the SPD constraint on the weight matrix $W$, a parametrization based on the Cholesky decomposition is used, as described in the \href{https://jbleger.gitlab.io/parametrization-cookbook}{Parametrization Cookbook} \citep{leger_parametrization_2023}.
Any SPD matrix $W$ can be uniquely decomposed as $W = LL^T$, where $L$ is a lower-triangular matrix with strictly positive diagonal entries.
Instead of optimizing $W$ directly under constraints, an unconstrained matrix $A$ is optimized.
$L$ is then constructed from $A$ by taking the lower-triangular part of $A$ and ensuring its diagonal entries are positive by applying the softplus function.
Specifically, $L_{ij} = A_{ij}$ for $i > j$, $L_{ii} = \text{softplus}(A_{ii}) = \log(1 + e^{A_{ii}})$, and $L_{ij} = 0$ for $i < j$.
The resulting matrix $W = LL^T$ is guaranteed to be SPD.
This construction is differentiable so $A$ can be optimized even if $W$ is used in practice.
$A$ is implemented as an \verb+nn.Linear+ module from the PyTorch library \citep{ansel_pytorch_2024}.

The \verb+AdamW+ optimizer \citep{loshchilov_decoupled_2017} is used with a learning rate of 0.1 and no weight decay.
Training stops when the optimized Spearman has not increased for 50 steps or after a maximum of 1000 steps.

For numerical stability the weight matrix $W$ is normalized to have Frobenius norm 1 at each step, and online min–max scaling is applied to neural distances $D^N$ so they remain in [0, 1].
These normalizations do not change the optimization problem as Spearman correlation is invariant to scaling.

\section{Batch Size Selection}
\label{sec:batch_size_selection}

To select an appropriate batch size for stochastic gradient descent, a preprocessing procedure is run on the feature values to estimate {\em correlation variability}.
The goal is to find a batch size that is small enough for computational efficiency but large enough to provide a stable estimate of the correlations between feature distances.
Formally, $p=64$ batches are sampled: $B_1,...,B_p$.
Starting with a small batch size $b=4096$, each batch $B_r$ is a set of $b$ pairs of indices:
$$B_r = \{(i_1, j_1), ..., (i_b, j_b)\}$$
The corresponding feature (and interaction) distances are computed and correlations between them are evaluated within the different batches.
Then, if the individual variabilities of those correlations across the $p$ batches are not too high (e.g., standard deviation below 0.01), this batch size is kept; otherwise it is increased by multiplying by a factor $\gamma=1.2$ ($b \leftarrow b \times \gamma$) and the estimation is restarted.
In practice, this preprocessing step can be unbounded for imbalanced datasets with very sparse features and relaxing the threshold may be necessary.

\section{Permutation Feature Importance Details}
\label{sec:pfi_details}

A trained MLEM model is defined by a matrix $W\in\mathbb{S}^{++}_m$ that can be used to predict neural distances $D_{ij}^N$ from feature distances $D_{ij}^F$ as follows:
$$D_{ij}^N \ \sim \ \widehat{D_{ij}^N} = \|D_{ij}^F\|_W = \sqrt{\sum_{1\leq k,l\leq m} W_{kl}D_{ij}^kD_{ij}^l}$$
To obtain feature importance for the interactions between features (corresponding to off-diagonal $W_{kl}$), the input space of the MLEM is reshaped to fit the Permutation Feature Importance framework.
Instead of considering the \mbox{$m$-dimensional} vector $D_{ij}^F$, it is transformed to an \mbox{$m_p$-dimensional} vector $P_{ij}$ where $m_p=\frac{m(m+1)}{2}$ is the number of pairs of features (including pairs ($f$, $f$) corresponding to a single feature).
For an index $1 \leq k' \leq m_p$ corresponding to a pair of features $(k,l)$ with $k\le l$ define:
\begin{equation*}
    P_{ij}^{k'}=
    \begin{cases}
        D_{ij}^{k}D_{ij}^{l} & \text{if } k = l \\
        2D_{ij}^{k}D_{ij}^{l} & \text{if } k \neq l
    \end{cases}
\end{equation*}
The factor 2 accounts for symmetry and avoids duplication.
Then the MLEM model can be rewritten as a mapping $h_W:\mathbb{R}^{m_p}\to \mathbb{R}$:
$$h_W(P_{ij}) = \sqrt{\sum_{1\leq k' \leq m_p} W_{kl}\,P_{ij}^{k'}}$$
With the same correspondence between $k'$ and $(k,l)$ as above.
Considering different pairs of stimuli as samples, Permutation Feature Importance can be applied to $h_W$ with Spearman correlation as the score to obtain importance for each feature and for each interaction between them.

\clearpage
\section{Weights}

\begin{figure}[!th]
    \centering
    \includegraphics[width=\textwidth]{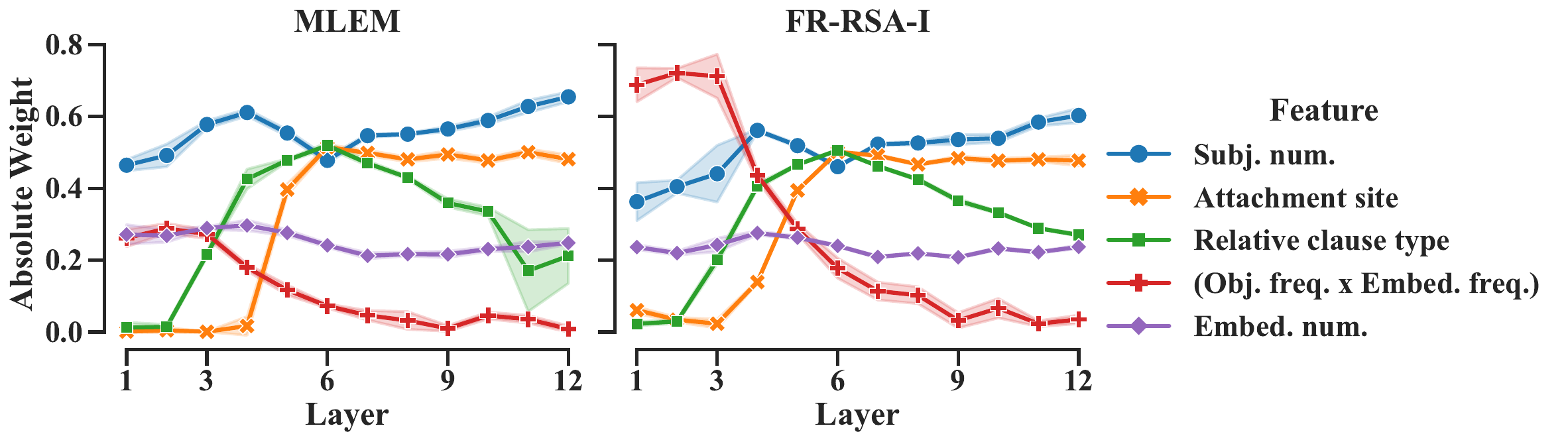}
    \caption{
        \textbf{Weight profiles are similar across methods but can be misleading.}
        This figure shows the absolute values of the learned weights for top features in FR-RSA-I (left) and MLEM (right) across BERT's 12 layers.
        While the profiles are broadly similar, raw weights are not always a reliable indicator of feature importance.
        For example, the interaction between "object frequency" and "embedded frequency" receives a large weight from both methods, but its permutation importance is negligible.
        This suggests that large weights can arise from model-internal parameter adjustments rather than true explanatory power.
    }
    \label{fig:weights}
\end{figure}

In addition to permutation importances, the absolute value of learned weights is inspected to understand the model's parameterization.
\cref{fig:weights} reports absolute weights for the top features across BERT layers for MLEM and FR-RSA-I.
While weight profiles are broadly similar across methods and layers, raw weights can be misleading.
A large absolute weight does not necessarily imply that a feature explains the geometry; it may reflect parameter scaling or compensate for other learned interactions.
For instance, the interaction between object frequency and embedded frequency has a large absolute weight for both methods but negligible importance.

\begin{figure}[!th]
    \centering
    \includegraphics[width=\textwidth]{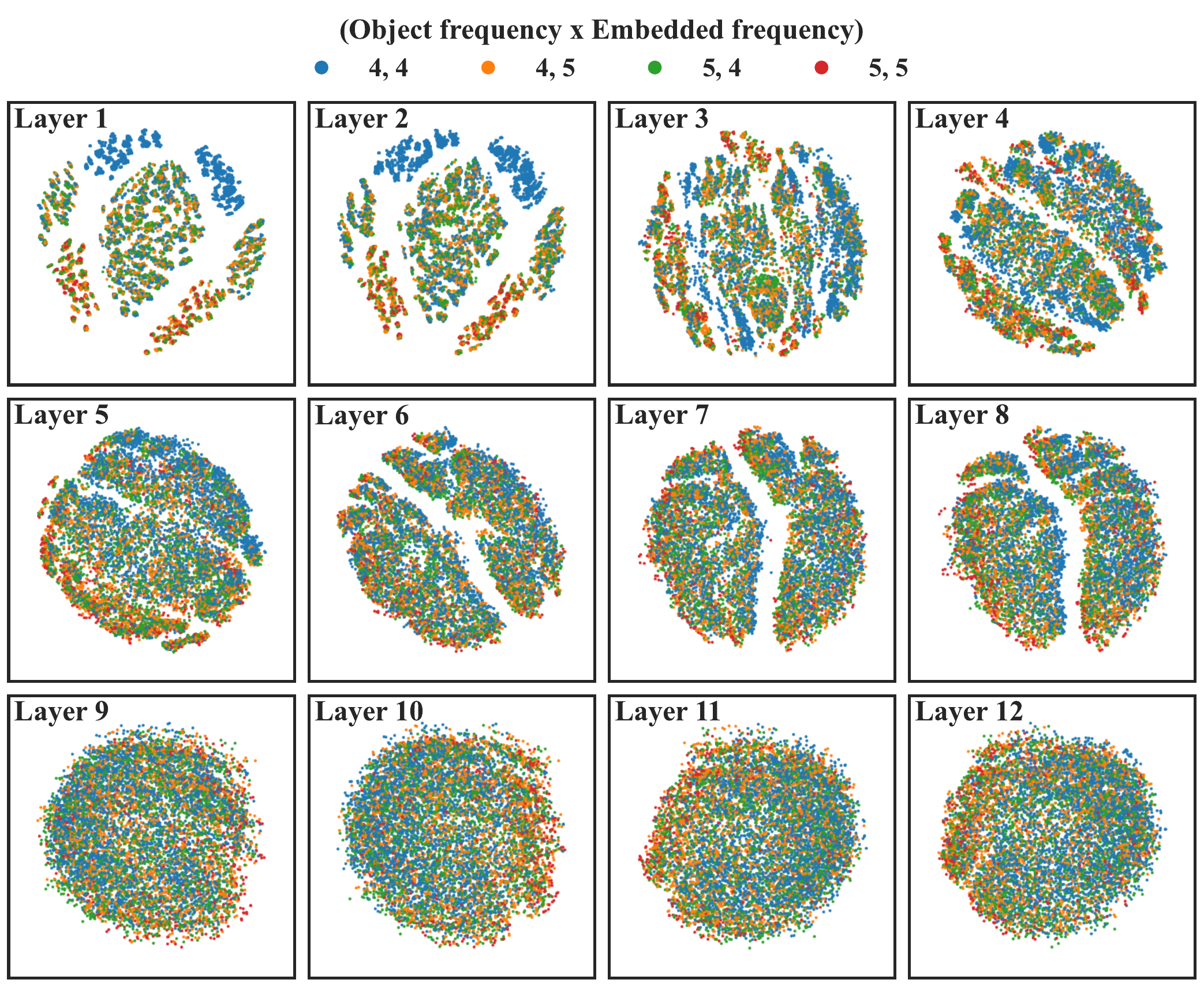}
    \caption{
        \textbf{Large weights do not always correspond to clear geometric structure.}
        This figure shows 2D MDS visualizations of BERT's sentence representations, colored by the interaction of "object frequency" and "embedded frequency".
        The clustering with respect to this interaction in early layers is not as clear as its weight seemed to suggest in \cref{fig:weights}, in particular by FR-RSA-I.
        This provides a concrete example of why permutation feature importance is a more reliable measure than raw weights for interpreting the geometry of neural representations.
    }
    \label{fig:mds_layers_false_positive}
\end{figure}

The MDS visualization in \cref{fig:mds_layers_false_positive} provides an example where weight magnitudes suggested a strong interaction, but the geometry did not clearly reflect the purported clustering.
This discrepancy suggests that feature importance is a more reliable measure than weights for understanding the geometry of neural representations.

\clearpage
\section{Quantitative comparison of methods}

To provide a more rigorous comparison between methods, we introduce quantitative measures of similarity for both weight and feature importance profiles.
These measures can be used to compare results from different methods (MLEM vs. FR-RSA-I), different models, or under varying conditions like noise level.

For weights, we use the Frobenius distance between two weight matrices.
This is used to compare estimated weights to ground truth in simulations (\cref{fig:fro_ground_truth}) and to assess robustness to noise (\cref{fig:fro_noiseless}).

For feature importance profiles, standard correlation measures like Spearman's $\rho$ are suboptimal as they treat all features equally.
The agreement on top-ranking features is of particular interest, as these have the most explanatory power.
Misranking features with low importance is less critical than misranking features with high importance.

Kendall's weighted $\tau$ \citep{vigna_weighted_2015} is a rank correlation coefficient that addresses this by assigning more weight to disagreements involving higher-ranked items.
Given $n$ items (features in our case) and two lists of scores for those items (feature importances in our case), $R$ and $S$, it is defined as:
$$\tau_w(R, S) = \frac{\sum_{i<j} w(i,j) \text{sgn}(R_i - R_j) \text{sgn}(S_i - S_j)}{\sum_{i<j} w(i,j)}$$
where the weight $w(i,j)$ is a function of the ranks of items $i$ and $j$.
A classical choice is hyperbolic weighting \citep{vigna_weighted_2015} which we describe next.
Let $\rho_R(i)$ be the rank of item $i$ in list $R$ (starting from 0).
The weight for a pair of items $(i,j)$ can be based on their ranks in list $R$:
$$w_R(i, j) = \frac{1}{\rho_R(i) + 1} + \frac{1}{\rho_R(j) + 1}$$
Similarly, a weight function $w_S$ can be defined based on the ranks in $S$.
To ensure symmetry, the final correlation $\tau_h$ is the average of the correlations computed with each weighting scheme:
$$\tau_h(R, S) = \frac{\tau_{w_R}(R, S) + \tau_{w_S}(R, S)}{2}$$
This symmetric approach ensures the measure does not depend on an arbitrary choice of which ranking to use for weights.
In practice, \verb+weightedtau+ from the \verb+scipy+ library \citep{virtanen_scipy_2020} is used, which implements this hyperbolic weighting scheme by default.

\subsection{Robustness to noise}

\begin{figure}[!th]
    \centering
    \includegraphics[width=\textwidth]{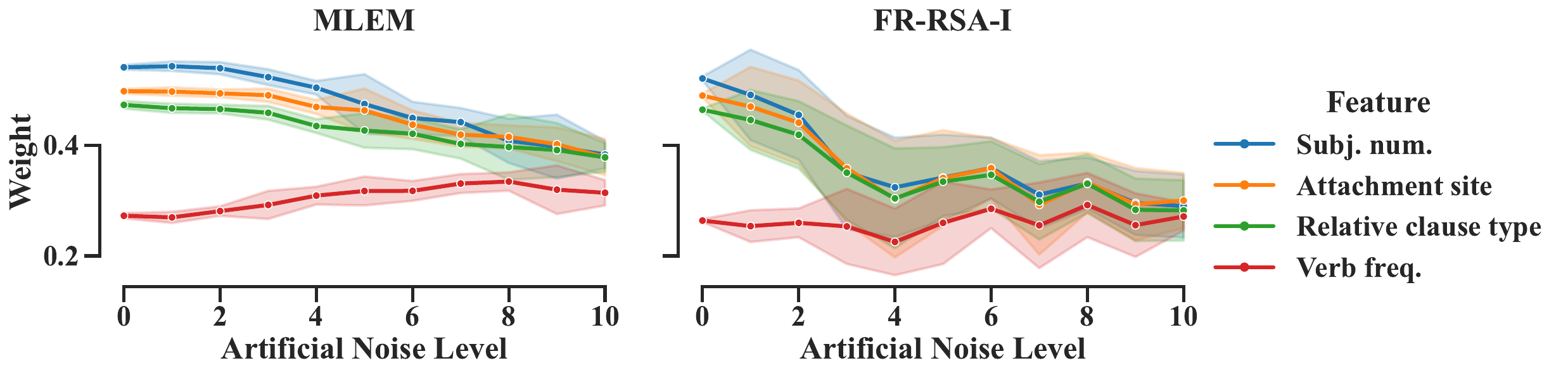}
    \caption{
        \textbf{MLEM's weight profile is more stable under noise than FR-RSA-I's.}
        The plots show the learned weights for top features in BERT's 7th layer for MLEM (left) and FR-RSA-I (right) under increasing levels of artificial noise.
        While both methods start with similar profiles at zero noise, the weight profile of FR-RSA-I degrades more rapidly as noise increases, showing greater instability.
        MLEM's profile remains more consistent, highlighting its superior robustness.
    }
    \label{fig:weights_noise}
\end{figure}

\begin{figure}[!th]
    \centering
    \includegraphics[width=\textwidth]{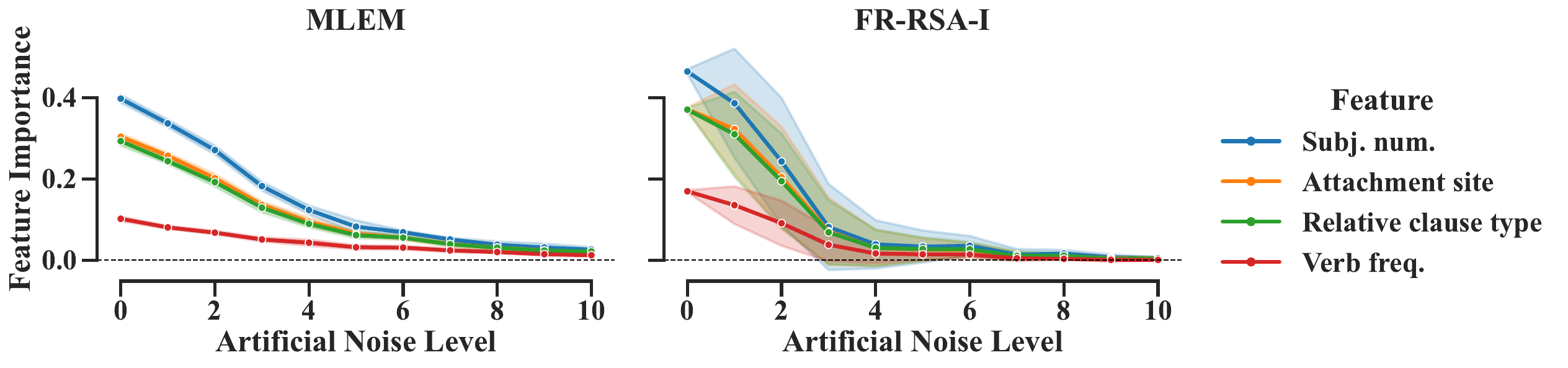}
    \caption{
        \textbf{MLEM's feature importance profile is more robust to noise.}
        This figure compares the feature importance profiles for FR-RSA-I (left) and MLEM (right) on BERT's 7th layer as artificial noise is added.
        Similar to the weight profiles, the feature importance profile for FR-RSA-I shows degradation with increasing noise.
        In contrast, MLEM's feature importance profile remains relatively stable, indicating that its interpretations are more reliable in noisy conditions.
    }
    \label{fig:fi_noise}
\end{figure}

\begin{figure}[!th]
    \centering
    \includegraphics[width=.5\textwidth]{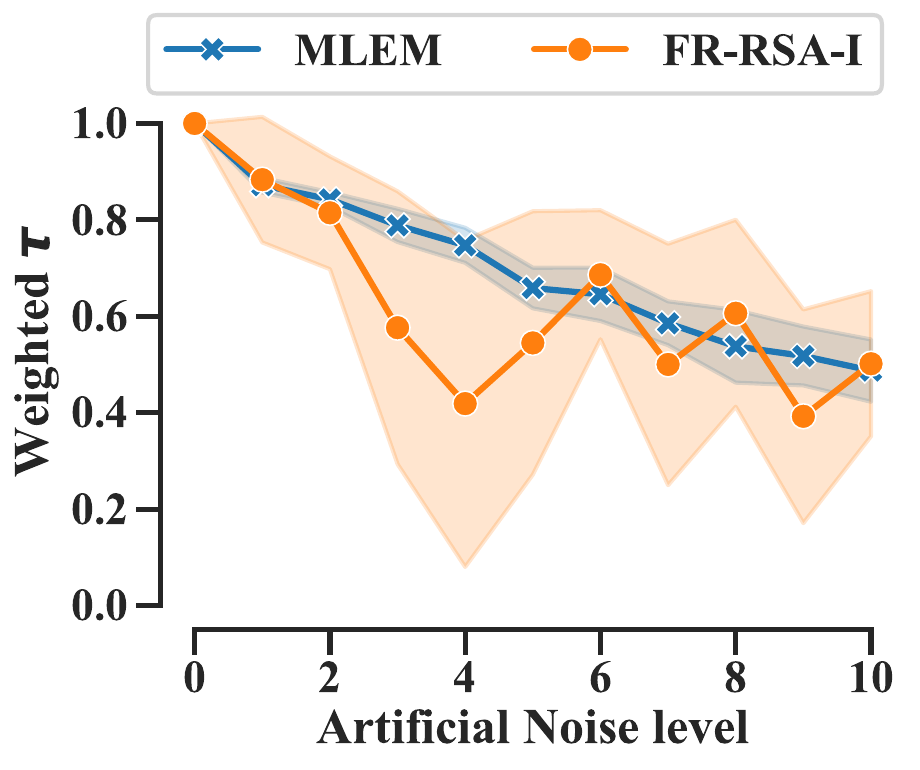}
    \caption{
        \textbf{MLEM's feature importance rankings are more stable against noise.}
        This plot shows Kendall's weighted $\tau$ correlation between the feature importance profile at each noise level and the noiseless feature importance profile, for MLEM (blue) and FR-RSA-I (orange).
        A higher correlation indicates greater stability.
        MLEM maintains a higher and more stable correlation as noise increases, whereas FR-RSA-I's correlation drops and shows high variability.
        This quantifies MLEM's superior robustness in preserving feature importance rankings under noise.
    }
    \label{fig:weighted_tau_noiseless}
\end{figure}

\clearpage
\subsection{Comparing MLEM and FR-RSA-I}

To further compare MLEM and FR-RSA-I, the Frobenius distance between their weight profiles and the Kendall's weighted $\tau$ between their feature importance profiles are computed across all BERT layers and noise levels.

\begin{figure}[!th]
    \centering
    \begin{subfigure}[b]{0.49\textwidth}
        \includegraphics[width=\textwidth]{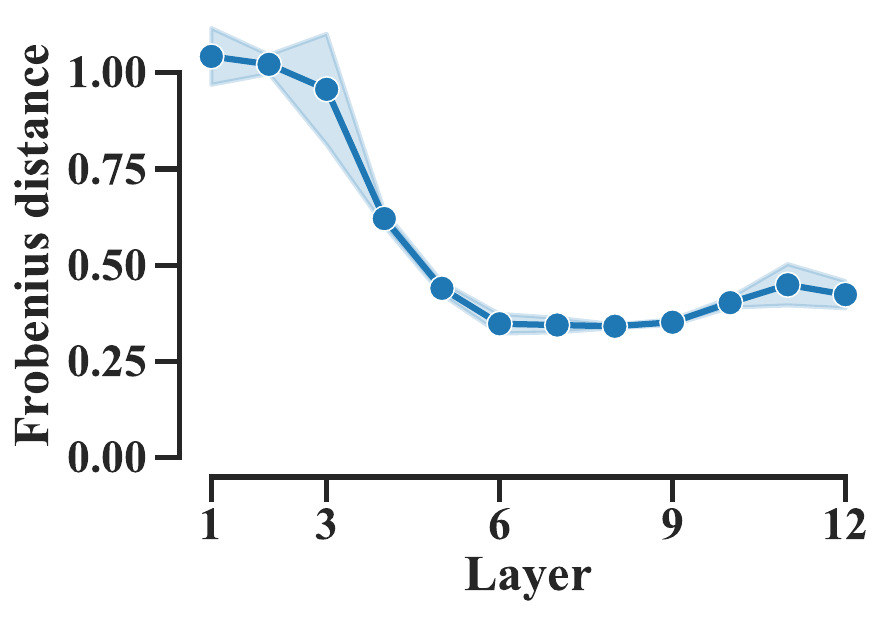}
        \caption{Dissimilarity of weight matrices.}
        \label{fig:fro_mlem_frrsa}
    \end{subfigure}
    \hfill
    \begin{subfigure}[b]{0.49\textwidth}
        \includegraphics[width=\textwidth]{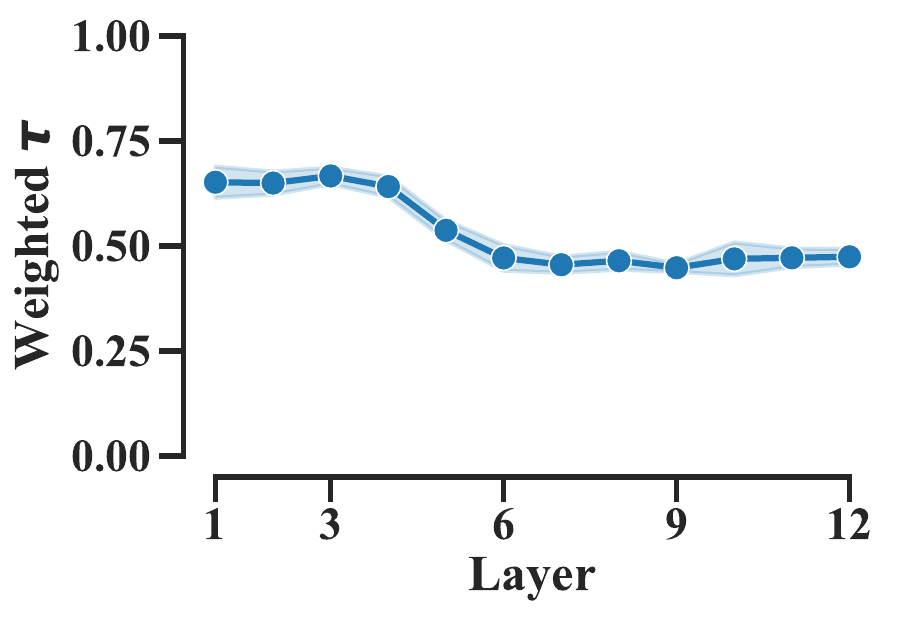}
        \caption{Agreement on feature importance.}
        \label{fig:weighted_tau_mlem_frrsa}
    \end{subfigure}
    \caption{
        \textbf{Comparison of MLEM and FR-RSA-I across BERT layers.}
        \textbf{(a)} The Frobenius distance between the weight matrices learned by MLEM and FR-RSA-I decreases in later BERT layers, suggesting the models converge to more similar weight solutions.
        \textbf{(b)} In contrast, the agreement on feature importance (Kendall's weighted $\tau$) is highest in early layers and decreases thereafter. This suggests that even when raw parameters become more similar, their interpretations in terms of feature importance can diverge.
    }
    \label{fig:mlem_frrsa_layers}
\end{figure}

\cref{fig:fro_mlem_frrsa} shows that the Frobenius distance between the weight matrices of the two methods decreases in the middle and higher layers of BERT, suggesting they converge to more similar weight solutions in these layers.
Conversely, \cref{fig:weighted_tau_mlem_frrsa} shows that the correlation between their feature importance profiles is highest in the early layers and decreases in later layers.
This indicates that while the raw weights may become more similar, the resulting interpretations in terms of feature importance diverge.
Given the earlier finding that feature importance is a more reliable measure, this divergence is significant.

\begin{figure}[!th]
    \centering
    \begin{subfigure}[b]{0.49\textwidth}
        \includegraphics[width=\textwidth]{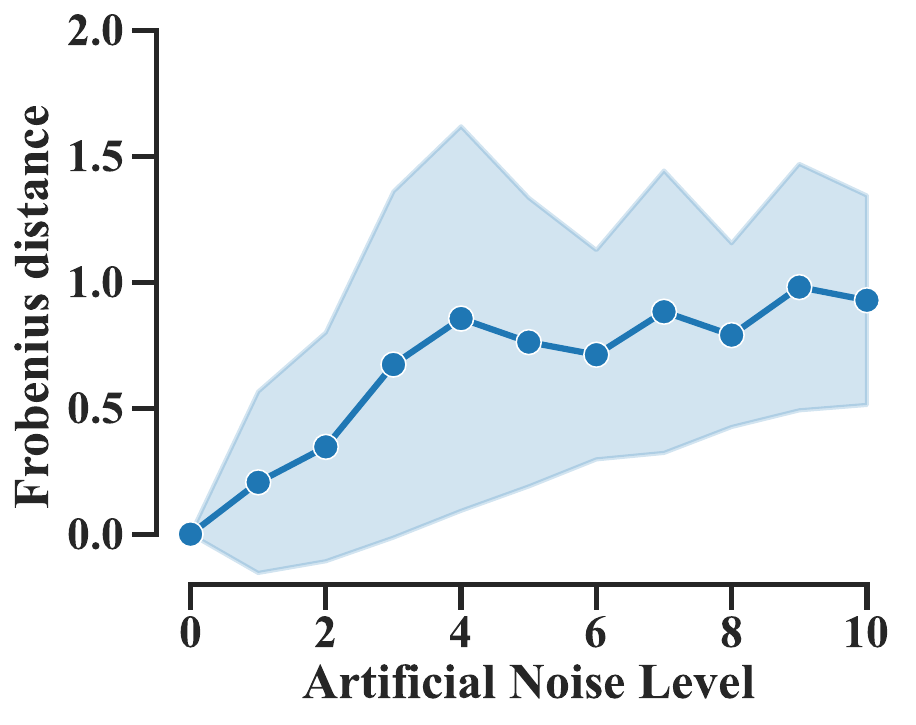}
        \caption{Dissimilarity of weight matrices.}
        \label{fig:fro_mlem_frrsa_noise}
    \end{subfigure}
    \hfill
    \begin{subfigure}[b]{0.49\textwidth}
        \includegraphics[width=\textwidth]{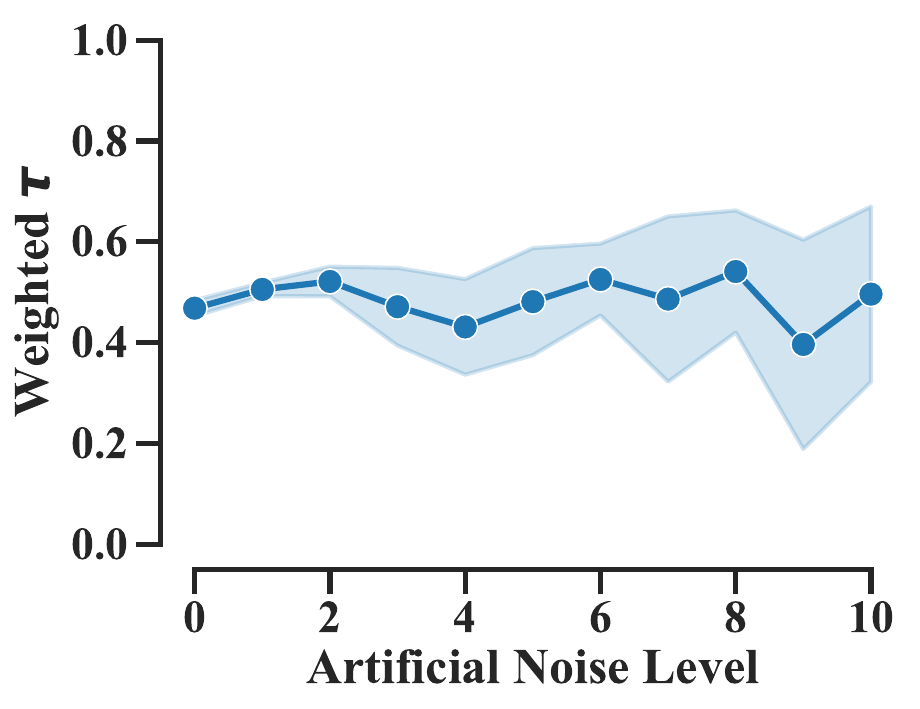}
        \caption{Agreement on feature importance.}
        \label{fig:weighted_tau_mlem_frrsa_noise}
    \end{subfigure}
    \caption{
        \textbf{Comparison of MLEM and FR-RSA-I under noise.}
        \textbf{(a)} The Frobenius distance between the weight matrices of MLEM and FR-RSA-I for BERT's 7th layer increases with noise, indicating they diverge more in noisier conditions.
        \textbf{(b)} Despite the growing dissimilarity in their weight matrices, the agreement on the ranking of important features (Kendall's weighted $\tau$) remains relatively stable, though with increased variability.
    }
    \label{fig:mlem_frrsa_noise}
\end{figure}

When examining the effect of noise (\cref{fig:fro_mlem_frrsa_noise} and \cref{fig:weighted_tau_mlem_frrsa_noise}), the weight matrices of the two methods diverge as noise increases, as expected.
However, the agreement on feature importances remains relatively stable, albeit with increased variability.

\clearpage
\subsection{MLEM and FR-RSA-I obtain the same encoding performance}

\begin{figure}[!th]
    \centering
    \begin{subfigure}[b]{0.32\textwidth}
        \includegraphics[width=\textwidth]{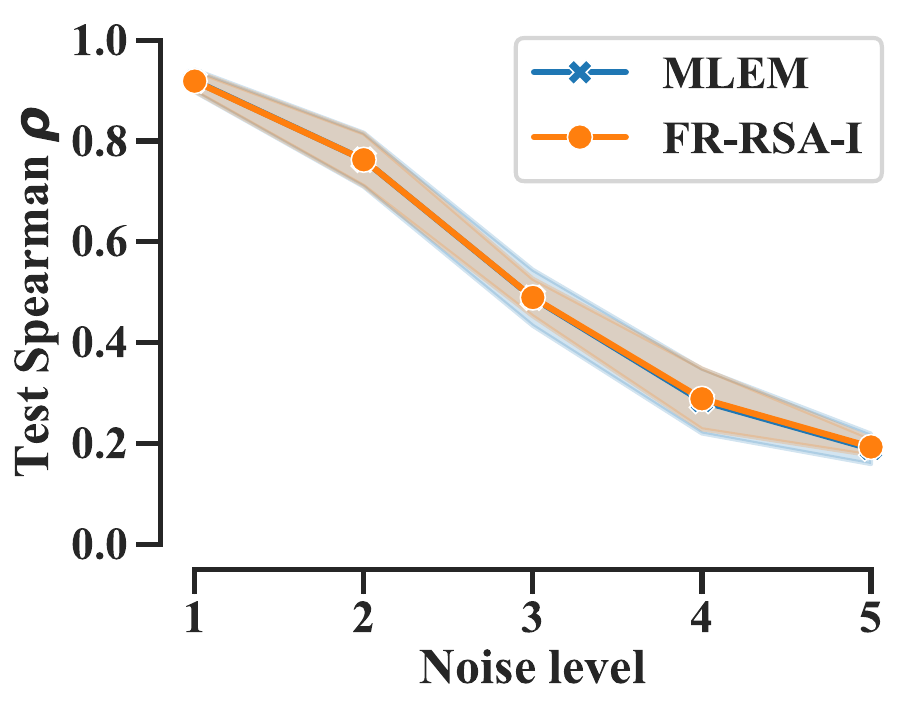}
        \caption{Simulated data.}
        \label{fig:spearman_simulation}
    \end{subfigure}
    \hfill
    \begin{subfigure}[b]{0.32\textwidth}
        \includegraphics[width=\textwidth]{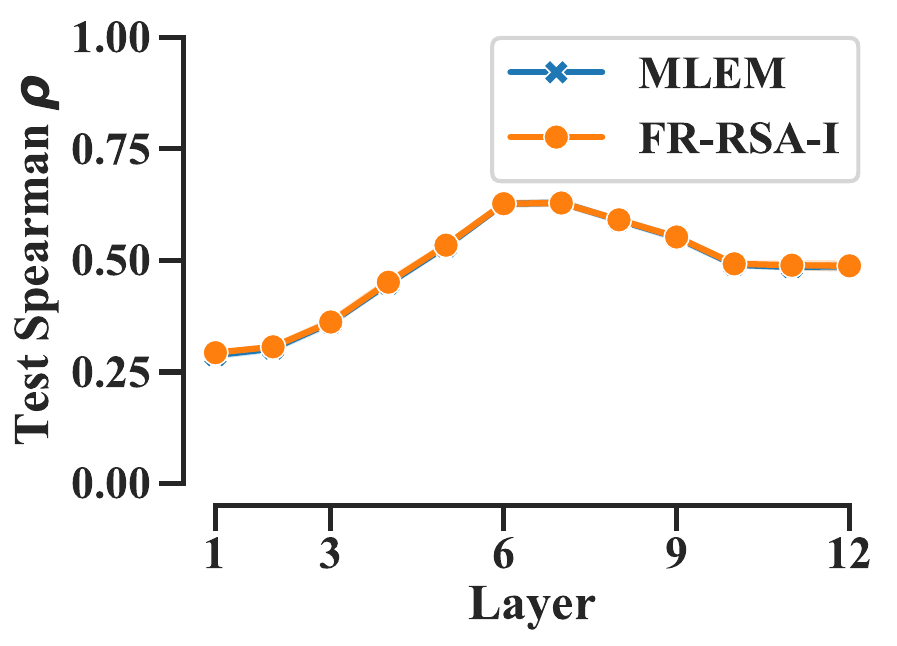}
        \caption{BERT layers.}
        \label{fig:spearman_layers}
    \end{subfigure}
    \hfill
    \begin{subfigure}[b]{0.32\textwidth}
        \includegraphics[width=\textwidth]{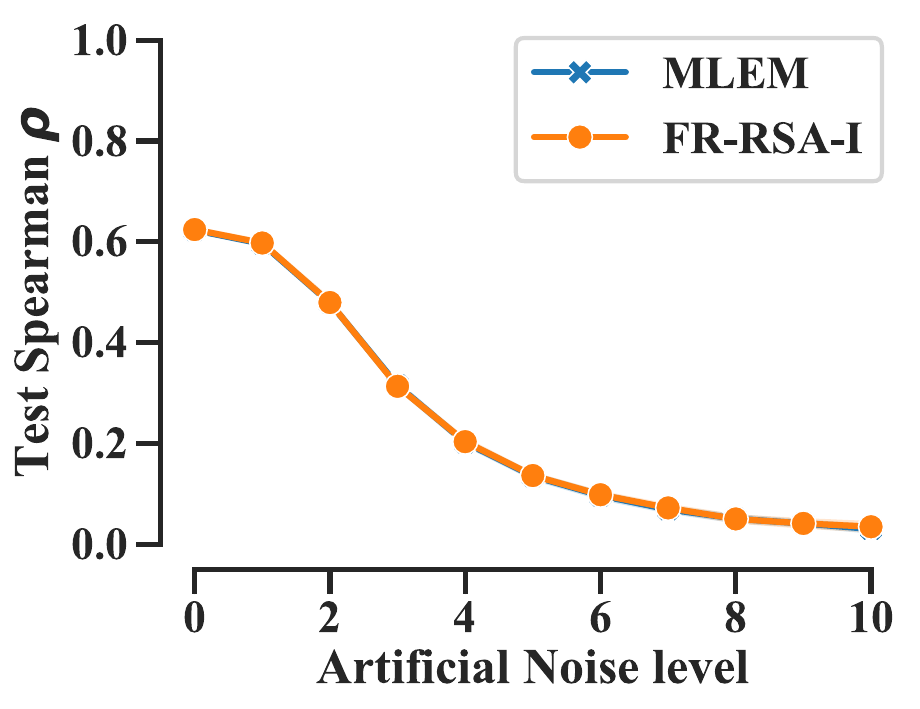}
        \caption{Artificial noise.}
        \label{fig:spearman_noise}
    \end{subfigure}
    \caption{
        \textbf{MLEM and FR-RSA-I achieve similar encoding performance.}
        The plots show the test Spearman correlation for MLEM (blue) and FR-RSA-I (orange).
        \textbf{(a)} On simulated data, both methods achieve comparable performance, which degrades as noise increases.
        \textbf{(b)} Across BERT layers, both models show nearly identical predictive performance.
        Performance peaks in middle layers, consistent with prior work \citep{hewitt_structural_2019}.
        \textbf{(c)} On BERT's 7th layer with artificial noise, performance degrades similarly for both methods.
        This confirms that MLEM's advantages do not come at the cost of encoding performance.
    }
    \label{fig:spearman_all}
\end{figure}

Although MLEM yields improved recovery of weight structure, improved robustness, and faster convergence, the two methods obtain similar final encoding performance measured by Spearman correlation on pairs of held-out stimuli (see \cref{fig:spearman_all}).
That is, both methods achieve comparable alignment to observed neural distances in terms of standard predictive scores.
The additional contributions of MLEM are therefore interpretability and stability: MLEM recovers more faithful feature interactions (by Frobenius to ground truth), yields more reliable feature importance profiles, and converges faster, while preserving encoding power.

\clearpage

\section{A decoding baseline is not informative}

\begin{figure}[!th]
    \centering
    \includegraphics[width=.8\textwidth]{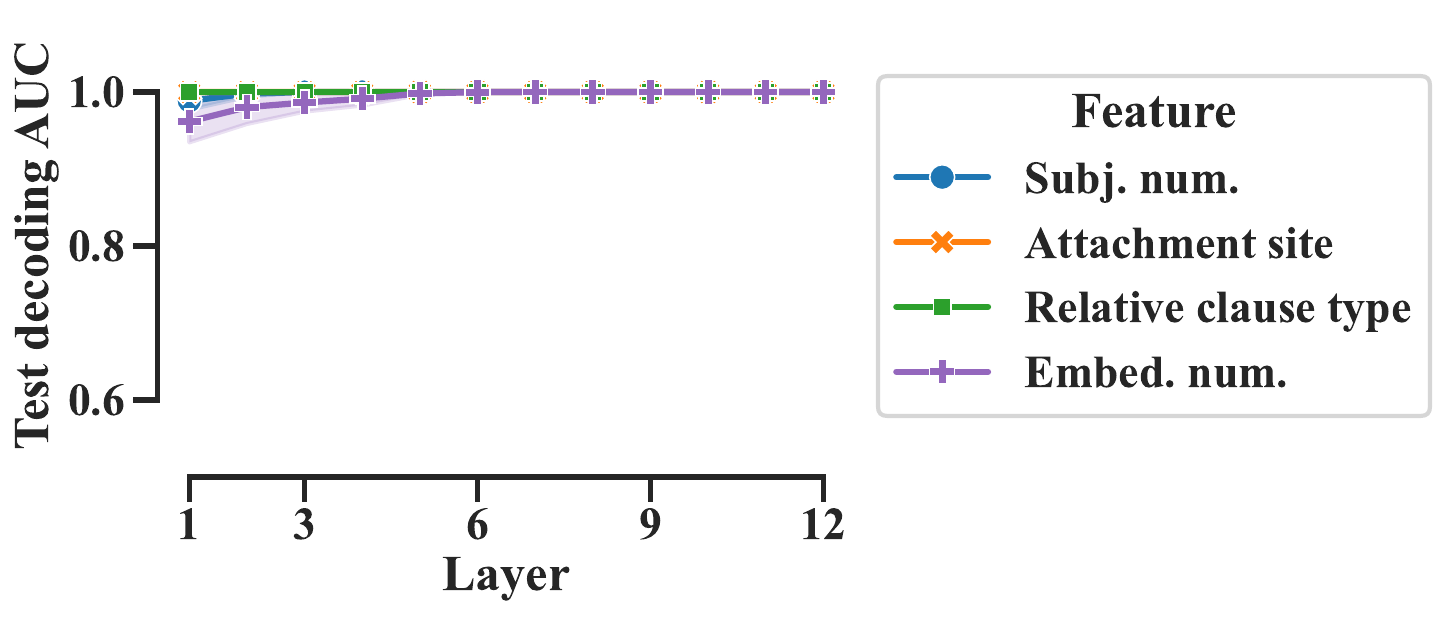}
    \caption{
        \textbf{Decoding accuracy provides limited insight into representational geometry.}
        This figure displays the decoding accuracy of selected linguistic features from BERT's layers using a linear classifier.
        Unlike the profiles of weights or feature importance, decoding accuracy remains consistently high across all layers, offering little differentiation of the underlying representational structure.
        The mean test AUC (full line) and standard deviation (shaded area) from 5-fold cross-validation are reported.
    }
    \label{fig:decoding}
\end{figure}

To contrast the encoding-based approach with a standard decoding baseline, linear classifiers are trained to decode linguistic features from BERT's layer-wise representations.
As shown in \cref{fig:decoding}, decoding accuracy for key features is high across all layers.
While this confirms that feature information is present, the flat accuracy profiles provide little insight into how the representational geometry changes across layers.
This highlights a key advantage of encoding models like MLEM, which are designed to reveal the structure of representations, not just the presence of information.

\clearpage

\section{A multivariate encoding baseline}
\label{sec:encoding_baseline}

\begin{figure}[!th]
    \centering
    \includegraphics[width=.8\textwidth]{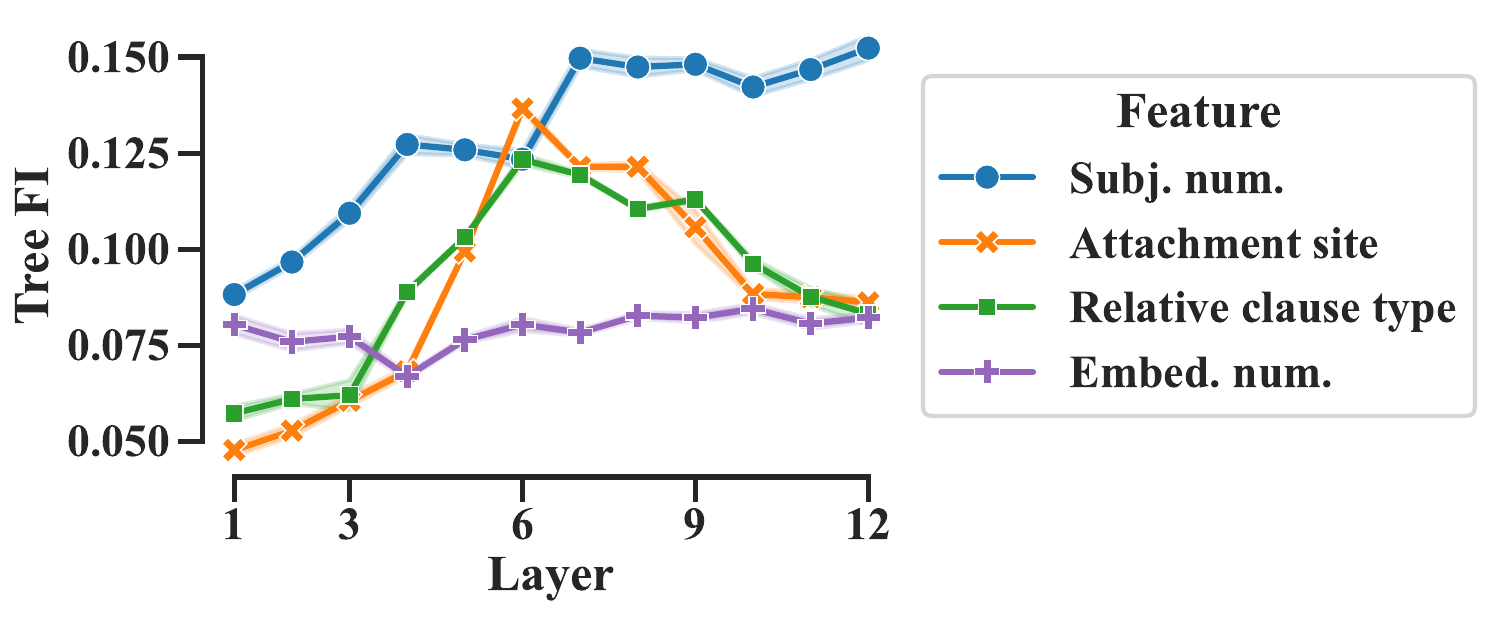}
    \caption{
        \textbf{A random forest encoding baseline recovers similar feature importance profiles but cannot model interactions.}
        This figure displays the feature importance from a random forest model trained to predict BERT's layer-wise representations from the theoretical features.
        The importance profile is similar to that found by MLEM, with syntactic features gaining importance in the middle layers.
        However, the model cannot capture feature interactions by design.
        The full line represents the average feature importance across the 100 trees of the forest, and the shaded area represents the standard deviation.
    }
    \label{fig:encoding_baseline}
\end{figure}

To further contextualize MLEM's performance, we compare it to a multivariate encoding baseline: a random forest regressor.
A random forest regressor was trained to predict the multi-dimensional BERT representations at each layer from the theoretical features.
Random forests can naturally handle multi-dimensional targets and categorical inputs.
We used the \verb+RandomForestRegressor+ implementation from \verb+scikit-learn+ and the built-in feature importance.
As shown in \cref{fig:encoding_baseline}, this baseline surprisingly recovers a feature importance profile very similar to that of MLEM and FR-RSA-I, with syntactic features peaking in the middle layers.
However, similarly to FR-RSA, standard random forests cannot model interactions between input features, hence the absence of an importance value for the interaction term. Furthermore, such an approach does not benefit from the noise and overfitting robustness conferred by the distance-based and Spearman-optimized framework of MLEM.

\clearpage

\section{Univariate vs. Multivariate MLEM}

\begin{figure}[!th]
    \centering
    \includegraphics[width=.8\textwidth]{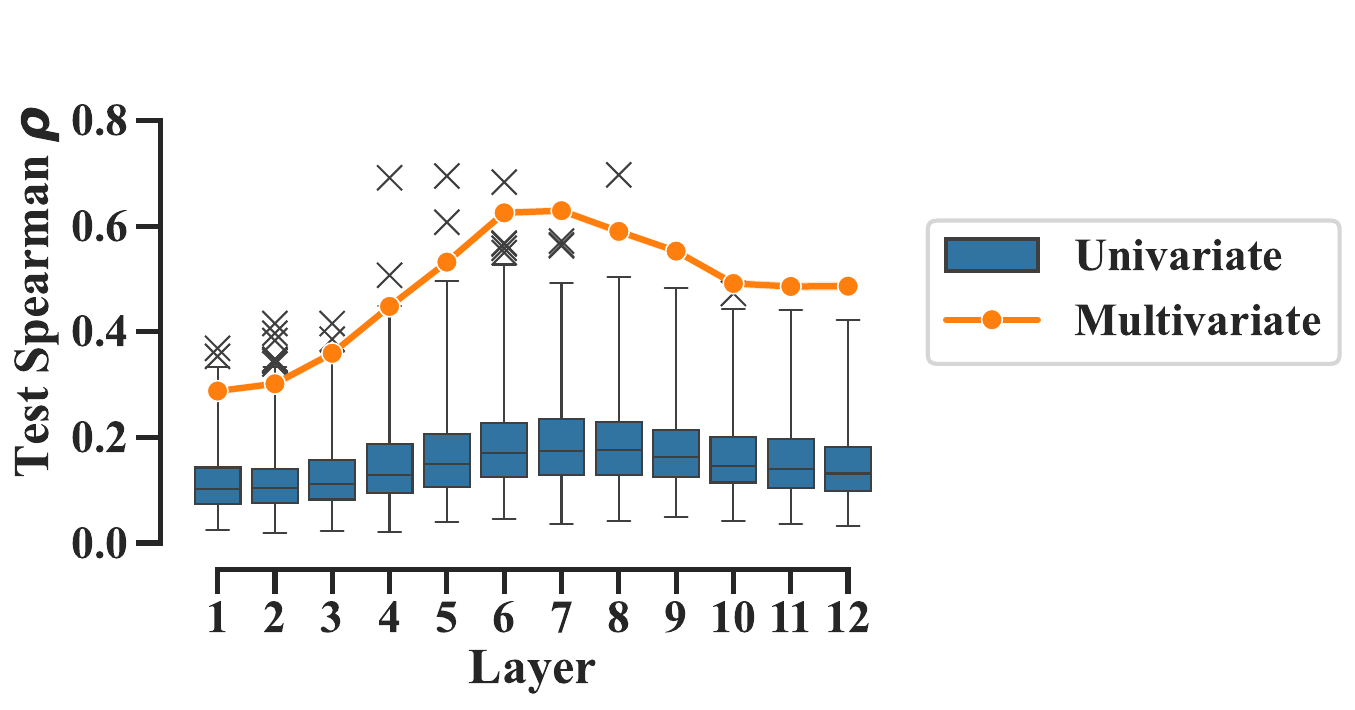}
    \caption{
        \textbf{Multivariate MLEM captures distributed representations more effectively than univariate models.}
        This figure compares the encoding performance (test Spearman correlation) of a single multivariate MLEM (orange) against the distribution of performances from univariate MLEMs (blue) for each layer of BERT.
        Each univariate model is trained on a single unit, while the multivariate model is trained on all the units in a layer.
        The superior performance of the multivariate MLEM indicates that it successfully captures geometric structure encoded across multiple units, which individual univariate models cannot.
        Units with a performance 3 times the interquartile range above the third quartile are considered outliers and are shown as crosses.
    }
    \label{fig:univariate}
\end{figure}

To assess whether MLEM captures distributed patterns of neural activity, the performance of the standard multivariate MLEM is compared against a univariate baseline.
The multivariate MLEM is trained on all the units of a given layer simultaneously, allowing it to learn weights that reflect interactions across units.
In contrast, a separate univariate MLEM is trained for each individual unit within a layer.
This univariate model can only capture information encoded in the activation of a single unit.
\cref{fig:univariate} compares the encoding performance (test Spearman correlation) of the multivariate MLEM with the distribution of performances from all univariate models for each layer of BERT.

\clearpage

\section{Other clusters of the BERT representations}

\begin{figure}[!th]
    \centering
    \includegraphics[width=.7\textwidth]{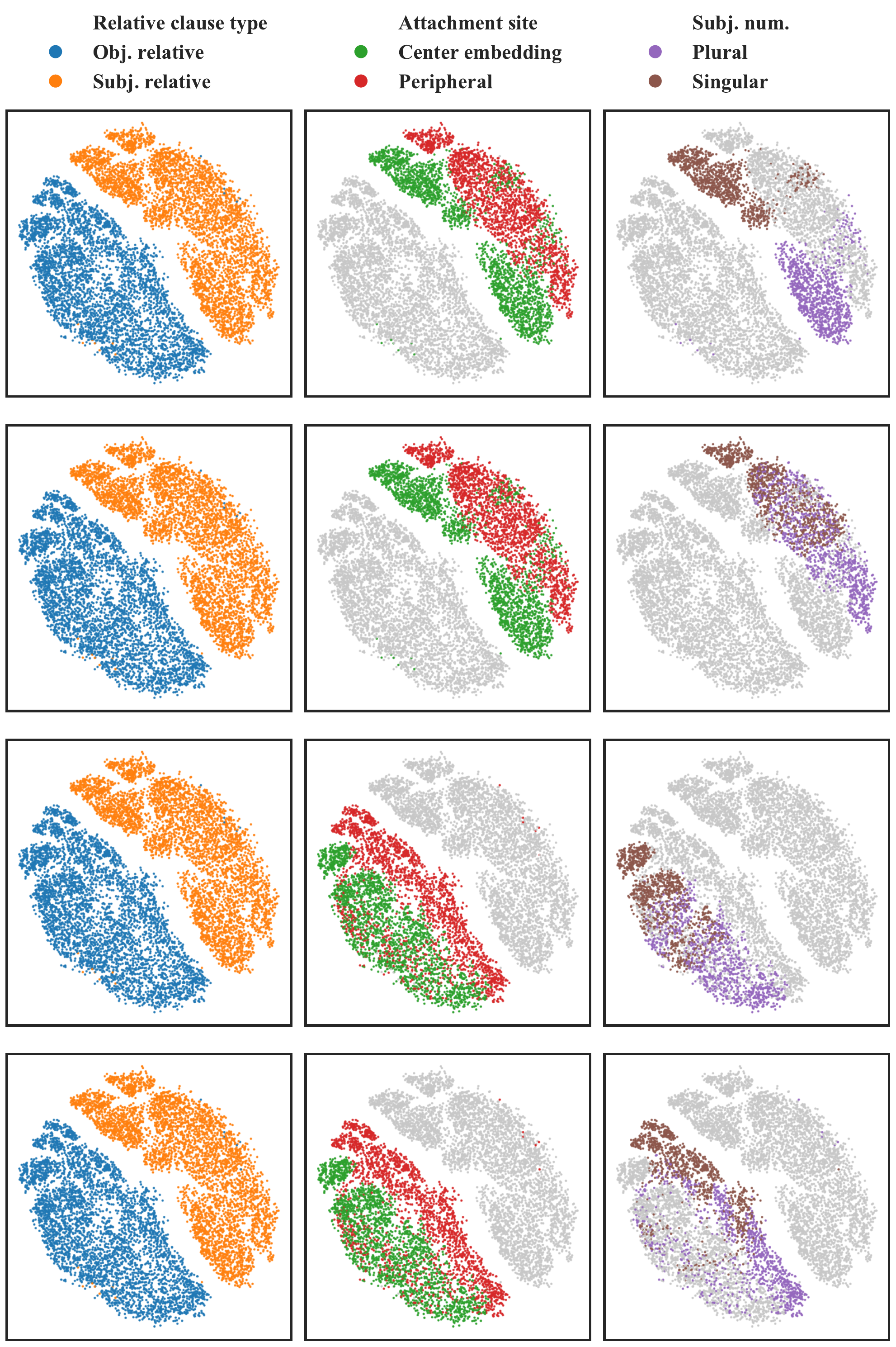}
    \caption{
        \textbf{Complete hierarchical clustering of BERT layer 6 representations.}
        This figure extends \cref{fig:hierarchical_mds} to show the full hierarchical clustering defined by the top three features from MLEM.
        Each row corresponds to a primary cluster based on "Relative Clause type" and "Attachment site".
        Within each of these, sub-clusters are formed based on "Subject number".
        This complete view illustrates how the combination of these three features partitions the entire representation space into a fine-grained, interpretable geometric structure.
    }
    \label{fig:hierarchical_mds_full}
\end{figure}

\cref{fig:hierarchical_mds} in the main text provides a walkthrough of the hierarchical clustering for a subset of the data.
For completeness, \cref{fig:hierarchical_mds_full} shows the full partitioning of the representation space for layer 6, illustrating how the top three features organize all sentence representations into a nested structure.

\clearpage

\section{Convergence speed}

\begin{figure}[!th]
    \centering
    \includegraphics[width=.6\textwidth]{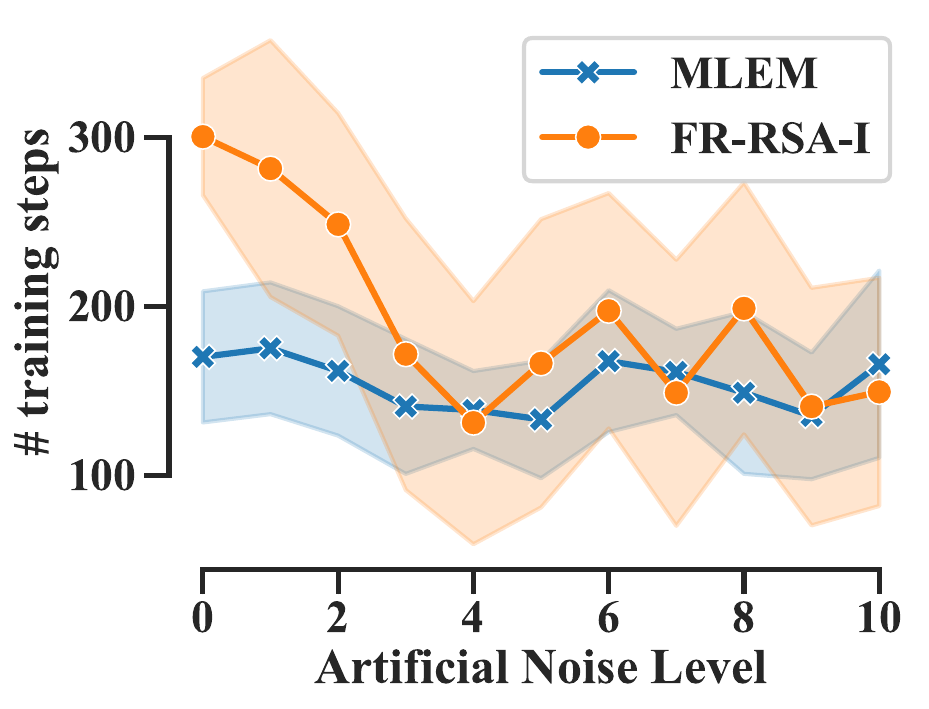}
    \caption{
        \textbf{MLEM's speed advantage diminishes at high artificial noise levels.}
        This plot shows the number of training steps for MLEM (blue) and FR-RSA-I (orange) on BERT's 7th layer with increasing artificial noise.
        While MLEM is faster at low noise levels, its speed advantage over FR-RSA-I decreases as noise increases, with both methods showing similar convergence times at high noise.
        However, MLEM's number of training steps exhibits less variability across runs.
    }
    \label{fig:n_epochs_noise}
\end{figure}